\documentclass{article}

\usepackage[final]{neurips_2023}

\usepackage{subfigure}
\usepackage{graphicx}

\usepackage[utf8]{inputenc} 
\usepackage[T1]{fontenc}    
\usepackage{hyperref}       
\usepackage{url}            
\usepackage{booktabs}       
\usepackage{amsfonts}       
\usepackage{nicefrac}       
\usepackage{microtype}      
\usepackage[dvipsnames]{xcolor}
\usepackage{amstext}
\usepackage{amsmath}
\usepackage{dsfont}
\usepackage{bbm}
\usepackage[capitalize,noabbrev,nameinlink]{cleveref}
\usepackage{multirow}
\usepackage{wrapfig}
\usepackage{float}
\usepackage{enumitem}
\usepackage[font=small,labelfont=bf]{caption}
\usepackage{siunitx}

\hypersetup{
    colorlinks=true,
    linkcolor=teal,
    filecolor=magenta,      
    urlcolor=ForestGreen,
    citecolor=NavyBlue,
    pdftitle={Tree-Rings Watermarks: Invisible Fingerprints for Diffusion Images},
    pdfpagemode=FullScreen,
    }
    
\title{Tree-Ring Watermarks: Fingerprints for Diffusion Images that are Invisible and Robust} 

\author{%
  Yuxin Wen, John Kirchenbauer, Jonas Geiping, Tom Goldstein \\
  University of Maryland \\
}

\begin{document}

\maketitle

\begin{abstract}
Watermarking the outputs of generative models is a crucial technique for tracing copyright and preventing potential harm from AI-generated content.
In this paper, we introduce a novel technique called \textit{Tree-Ring Watermarking} that robustly fingerprints diffusion model outputs.  Unlike existing methods that perform post-hoc modifications to images after sampling, \textit{Tree-Ring Watermarking} subtly influences the entire sampling process, resulting in a model fingerprint that is invisible to humans. The watermark embeds a pattern into the initial noise vector used for sampling. These patterns are structured in Fourier space so that they are invariant to convolutions, crops, dilations, flips, and rotations.  After image generation, the watermark signal is detected by inverting the diffusion process to retrieve the noise vector, which is then checked for the embedded signal.  We demonstrate that this technique can be easily applied to arbitrary diffusion models, including text-conditioned Stable Diffusion, as a plug-in with negligible loss in FID. Our watermark is semantically hidden in the image space and is far more robust than watermarking alternatives that are currently deployed. Code is available at \url{https://github.com/YuxinWenRick/tree-ring-watermark}.
\end{abstract}

\section{Introduction}

The development of diffusion models has led to a surge in image generation quality. Modern text-to-image diffusion models, like Stable Diffusion and Midjourney, are capable of generating a wide variety of novel images in an innumerable number of styles. These systems are general-purpose image generation tools, able to generate new art just as well as photo-realistic depictions of fake events for malicious purposes.

The potential abuse of text-to-image models motivates the development of  \textit{watermarks} for their outputs. A watermarked image is a generated image containing a signal that is invisible to humans and yet marks the image as machine-generated. Watermarks document the use of image generation systems, enabling social media, news organizations, and the diffusion platforms themselves to mitigate harms or cooperate with law enforcement by identifying the origin of an image~\citep{bender_dangers_2021,grinbaum_ethical_2022}.

Research and applications of watermarking for digital content have a long history, with many approaches being considered over the last decade \citep{oruanaidh_rotation_1997,langelaar_watermarking_2000}. However, so far research has always conceptualized the watermark as a minimal modification imprinted onto an existing image \citep{solachidis_circularly_2001,chang_svd-based_2005,liu_optimized_2019,fei_supervised_2022}. For example, the watermark currently deployed in Stable Diffusion \citep{10.5555/1564551}, works by modifying a specific Fourier frequency in the generated image.

The watermarking approach we propose in this work is conceptually different: This is the first watermark that is truly invisible, as no post-hoc modifications are made to the image.  Instead, the \textit{distribution of generated images is imperceptibly modified and an image is drawn from this modified distribution.} This way, the actual sample carries no watermark in the classical additive sense, however an algorithmic analysis of the image can detect the watermark with high accuracy. From a more practical perspective, the watermark materializes in minor changes in the potential layouts of generated scenes, that cannot be distinguished from other random samples by human inspection.

\looseness -1 This new approach to watermarking, which we call \textit{Tree-Ring Watermarking} based on the patterns imprinted into the Fourier space of the noise vector of the diffusion model, can be easily incorporated into existing diffusion model APIs and is invisible on a per-sample basis. Most importantly, \textit{Tree-Ring Watermarking} is \emph{far more robust than existing methods} against a large battery of common image transformations, such as crops, color jitter, dilation, flips, rotations, or noise. \textit{Tree-Ring Watermarking} requires no additional training or finetuning to implement, and the watermark can only be detected by parties in control of the image generation model. We validate the watermark in a number of tests, measuring negligible impact on image quality scores, high robustness to transformations, the low false-positive rate in detection, and usability for arbitrary diffusion models both with and without text conditioning.

\begin{figure}[t]
    \centering
    \includegraphics[width=0.95\textwidth]{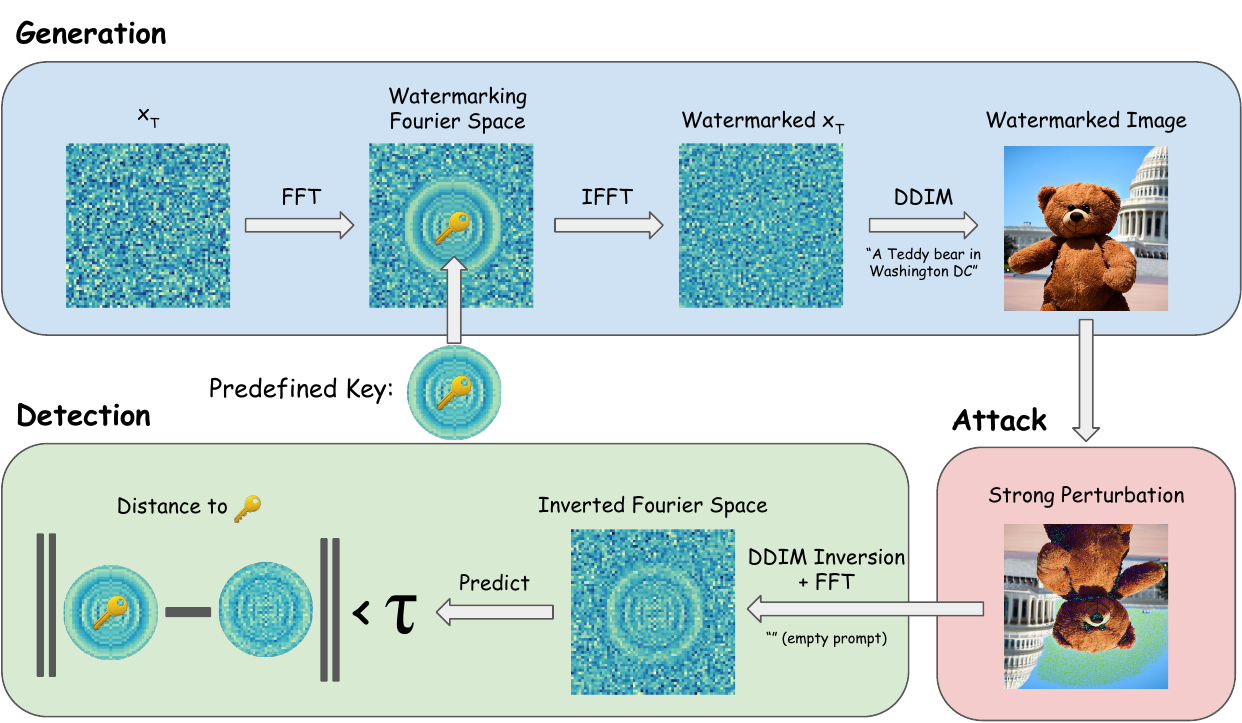}
    \caption{Pipeline for \textit{Tree-Ring Watermarking}. A diffusion model generation is watermarked and later detected through ring-patterns in the Fourier space of the initial noise vector.}
    \label{fig:teaser}
    \vspace{-.4cm}
\end{figure}

\section{Related Work}

\paragraph{Diffusion Models}
Diffusion Models, arising out of the score-based generative models of the formalism of \citet{song_generative_2019,song_improved_2020}, are the currently strongest models for image generation \citep{ho_denoising_2020,dhariwal_diffusion_2021}. Diffusion models are capable of sampling new images at inference time by iteratively processing an initial noise map. The most prominent sampling algorithm in deployment is DDIM sampling \citep{nichol_improved_2021} without additional noise, which can generate high-quality images in fewer steps than traditional DDPM sampling. Diffusion models are further accelerated for practical usage by optimizing images only in the latent space of a pre-trained VAE, such as in latent diffusion \citep{rombach_high-resolution_2022}.

\paragraph{Watermarking Digital Content}
Strategies to imprint watermarks onto digital content, especially images, have a long tradition in computer vision. Approaches such as \citet{boland_watermarking_1996,cox_secure_1996,oruanaidh_rotation_1997} describe traditional watermark casting strategies based on imprinting a watermark in a suitable frequency decomposition of the image, constructed through DCT, DWT, Fourier-Mellin, or complex wavelet transformations. These frequency transformations all share the beneficial property that simple image manipulations, such as translations, rotations, and resizing are easily understandable and watermarks can be constructed with robustness to these transformations in mind. A fair evaluation of watermarking approaches appears in \citet{pitas_method_1998}, \citet{kutter_fair_1999}, which highlight the importance of measurement of false-positive rates for each strategy and ROC-curves under attack through various image manipulations.
Work continues on imprinting watermarks, with strategies based on SVD decompositions \citep{chang_svd-based_2005}, Radon transformations \citep{seo_robust_2004} and based on multiple decompositions \citep{al-haj_combined_2007}.

\begin{figure}
    \centering
    \subfigure[W/o Watermark]{\includegraphics[width=0.49\textwidth]{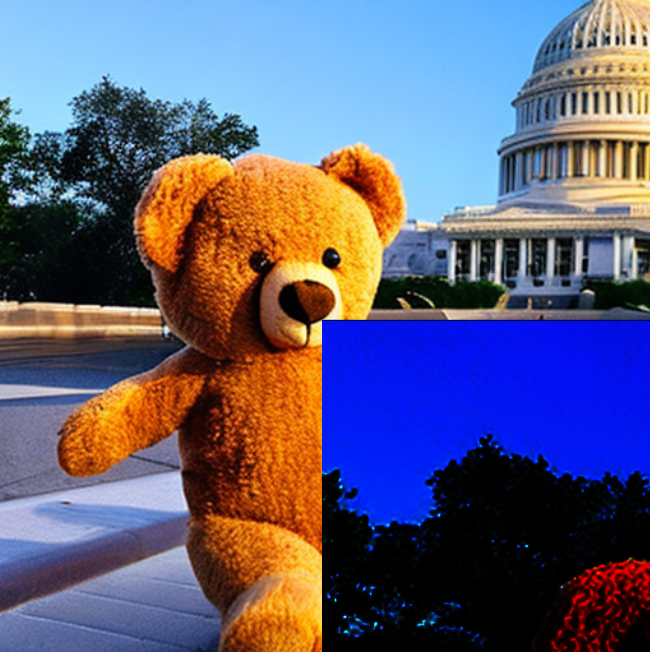}}
    \subfigure[DwtDct]{\includegraphics[width=0.49\textwidth]{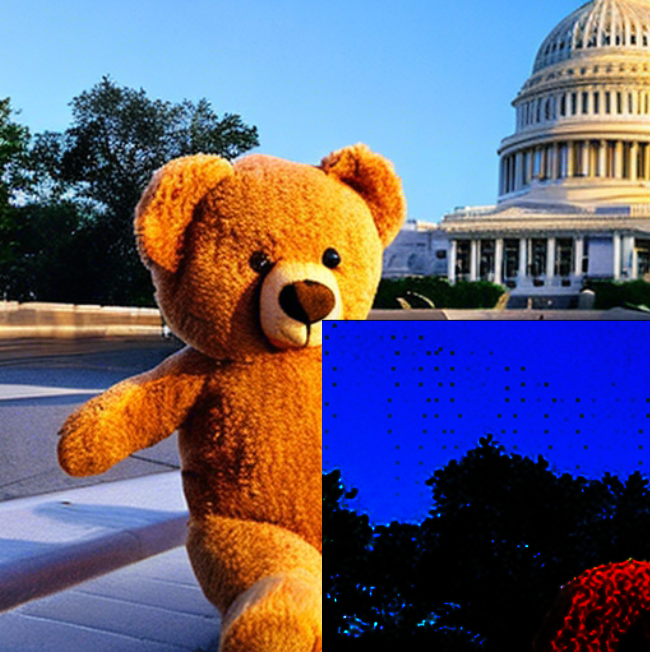}}
    \subfigure[RivaGAN]{\includegraphics[width=0.49\textwidth]{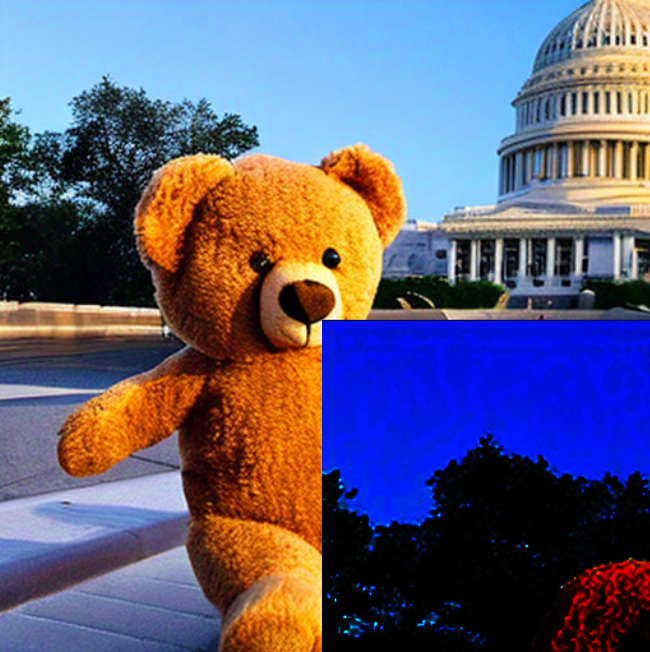}}
    \subfigure[\textit{Tree-Ring} (Ours)]{\includegraphics[width=0.49\textwidth]{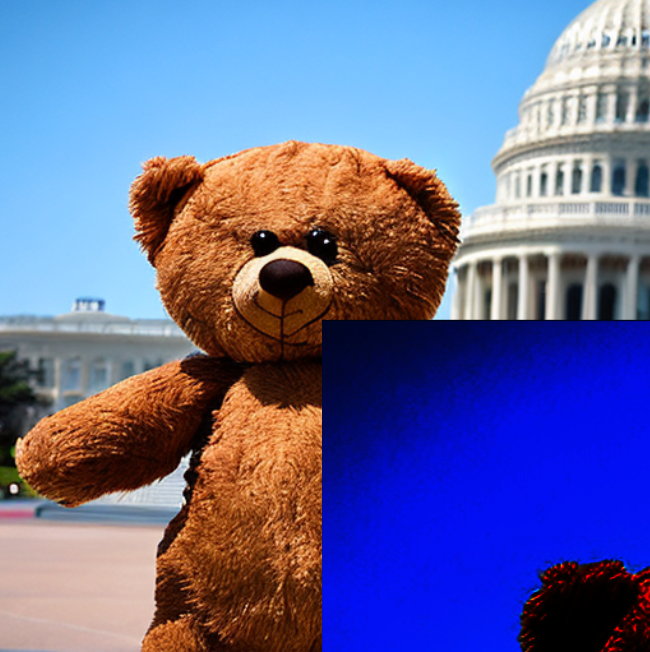}}
    \caption{Various watermarked generations with the same random seed are presented, showcasing the ``invisible'' nature of our proposed watermark. A zoomed-in view with high contrast is provided in the bottom right corner. For more high-resolution watermarked images, please refer to Supplementary Material.}
    \label{fig:related_work_comparison}
    \vspace{-.25cm}
\end{figure}

\paragraph{Fingerprinting and Watermarking Generative Models}
The development of modern deep neural networks opened up new possibilities for ``deep'' watermarking. \citet{hayes_generating_2017} and  \citet{zhu_hidden_2018} propose strategies to learn watermarking end-to-end, where both the watermark encoder and the watermark decoder are learned models, optimized via adversarial objectives to maximize transmission and robustness \citep{zhang_robust_2019}.  \citet{zeng2023securing} present a related approach, in which a neural network watermarked encoder and its associate detector are jointly learned using an image dataset.  
Notably these approaches still work like a traditional watermark in that the encoder imprints a post-hoc signal onto a given image - however the type of imprint is now learned. We refer to \citet{wan_comprehensive_2022} for an overview. A recent improvement is two-stage processes like \citet{yu_artificial_2022}, where the trained encoder is used to imprint the watermark onto the training data for a generative model. This leads to a trained generative model where the watermark encoder is ``baked in'' to the model, making it easier to generate watermarked data. The Stable Signature of \citet{fernandez_stable_2023}, applies this idea to latent diffusion models by finetuning the latent decoder based on a pre-trained watermark encoder. \citet{zhao_recipe_2023} similarly train on watermarked data for unconditional diffusion models.

Existing image watermarking approaches first learn a watermark signal and then learn to either embed it into generated data or the generating model. This pipeline stands in contrast to watermarking approaches for language models such as \citet{kirchenbauer_watermark_2023}. There, no training is necessary to generate watermarked data and the output distribution of the generative model is altered to encode a watermark into generated data in a distributional sense. In the same vein, we propose an approach to alter the output distribution of diffusion models to effectively watermark their outputs. As discussed, this has a number of advantages, in comparison to related work we especially highlight that no training is necessary, that the watermark works with existing models, and that this is the first watermark that does not rely on minor modification of generated images. In this sense, this is the first watermark that is really ``invisible'', see \cref{fig:related_work_comparison}.

We note in passing that watermarking the output of generative models is not to be confused with the task of watermarking the weights of whole models, such as in \citet{uchida_embedding_2017,zhang_protecting_2018,bansal_certified_2022}, who are concerned with identifying and fingerprinting models for intellectual property reasons.

\subsection{Diffusion Models and Diffusion Inversion} \label{inverse}
We first introduce the basic notation for diffusion models and DDIM sampling \citep{ho_denoising_2020, song_improved_2020, dhariwal_diffusion_2021}.
A forward diffusion process consists of $T$ steps of the noise process a predefined amount of Gaussian noise vector to a real data point $x_0 \in q(x)$, where $q(x)$ is the real data distribution, specifically:
\begin{equation*}
    q(x_t | x_{t-1}) = \mathcal{N}(x_t; \sqrt{1-\beta_t}x_t, \beta_t\mathbf{I}), \text{for } t \in \{0, 1, ..., T-1\},
\end{equation*}
where $\beta_t \in (0, 1)$ is the scheduled variance at step $t$. The closed-form for this sampling is 
\begin{equation}
    \label{equation:forward_diffusion}
    x_t = \sqrt{\Bar{\alpha}_t}x_0 + \sqrt{1 - \Bar{\alpha}_t}\epsilon,
\end{equation}
where, $\Bar{\alpha}_t = \prod_{i=0}^{t}(1-\beta_{t})$.

For the reverse diffusion process, DDIM \citep{song_improved_2020} is an efficient deterministic sampling strategy, mapping from a Gaussian vector $x_T \sim \mathcal{N}(0,\,1)$ to an image $x_0 \in q(x)$.
For each denoising step, a learned noise-predictor $\epsilon_{\theta}$ estimates the noise $\epsilon_{\theta}(x_t)$ added to $x_0$. According to \cref{equation:forward_diffusion}, we can derive the estimation of $x_0$ as:
\begin{equation*}
    \hat{x}_0^t = \frac{x_t-\sqrt{1-\Bar{\alpha}_t}\epsilon_{\theta}(x_t)}{\sqrt{\Bar{\alpha}_t}}.
\end{equation*}
Then, we add the estimated noise to $\hat{x}_0$ to find $x_{t-1}$:
\begin{equation*}
    x_{t-1} = \sqrt{\Bar{\alpha}_{t-1}}\hat{x}_0^t + \sqrt{1-\Bar{\alpha}_{t-1}}\epsilon_{\theta}(x_t).
\end{equation*}
We denote such a recursively denoising process from $x_T$ to $x_0$ as $x_0 = \mathcal{D}_\theta(x_T)$.

However, given the learned model $\epsilon_{\theta}(x_t)$, it is also possible to move in the opposite direction\footnote{\looseness -1 To reduce confusion we will always describe the generative diffusion process that goes from $x_T$ to $x_0$ as the ``reverse process''.  We use ``inverse process'' to denote the estimation of the noise vector $x_T$ from the final output $x_0$.}. Starting from an image $x_0$,  \citet{dhariwal_diffusion_2021} describes an inverse process that retrieves an initial noise vector $x_T$ which maps to an image $\hat{x}_0$ close to $x_0$ through DDIM, where $\hat{x}_0 = \mathcal{D}_\theta(x_T, 0) \approx x_0$. This inverse process depends on the assumption that $x_{t-1} - x_{t} \approx x_{t+1} - x_{t}$. Therefore, from $x_{t} \to x_{t+1}$, we follow:
\begin{equation*}
    x_{t+1} = \sqrt{\Bar{\alpha}_{t+1}}\hat{x}_0^t + \sqrt{1-\Bar{\alpha}_{t+1}}\epsilon_{\theta}(x_t).
\end{equation*}
We denote the whole inversion process from a starting real image $x_0$ to $x_T$ as $x_T = \mathcal{D}^\dagger_\theta(x_0)$.

In this work, we re-purpose DDIM inversion $\mathcal{D}^\dagger_\theta$ for watermark detection. Given a generated image $x_0$ with a starting noise $x_{T}$, we apply DDIM inversion to find $\hat{x}_T$. We empirically find DDIM's inversion performance to be quite strong, and $\hat{x}_T \approx x_{T}$. While it may not be surprising that inversion is accurate for unconditional diffusion models, inversion also succeeds well-enough for conditional diffusion models, even when the conditioning $c$ is not provided.  This property of inversion will be exploited heavily by our watermark below.

\section{Method}
In this section,
we provide a detailed description of each layer of \textit{Tree-Ring Watermarking}.

\subsection{Threat Model}
We first briefly describe the threat model considered in this work and clarify the setting:
The goal of watermarking is to allow for image generation without quality degradation while enabling the model owner the ability to identify if a given image is generated from their model. Meanwhile, the watermarked image is used in every-day applications and subject to a number of image manipulations and modifications. We formalize this as an adversary who tries to remove the watermark in the generated image to evade detection using common image manipulations, but note that informally, we are also interested in watermark robustness across common usage. Ultimately, this setup leads to a threat model with two agents that act sequentially. 

\begin{itemize}
\item Model Owner (Generation Phase): Gene owns a generative diffusion model $\epsilon_\theta$ and allows images $x$ to be generated through an API containing the private watermarking algorithm $\mathcal{T}$. The watermarking algorithm $\mathcal{T}$ should have a negligible effect on the generated distribution, so that quality is maintained and watermarking leaves no visible trace.
\item Forger: Fiona generates an image $x$ through the API, then tries to evade the detection of $\mathcal{T}$ by applying strong data augmentations that convert $x$ to $x'$. Later, Fiona uses $x'$ for a prohibited purpose and claims that $x'$ is her intellectual property.
\item \looseness -1 Model Owner (Detection Phase): Given access to $\epsilon_\theta$ and $\mathcal{T}$, Gene tries to determine if  $x'$ originated from $\epsilon_\theta$. Gene has no knowledge of the text used to condition the model, or other hyperparameters like guidance strength and the number of generation steps.
\end{itemize}

\subsection{Overview of \textit{Tree-Ring Watermarking}}
Diffusion models convert an array of Gaussian noise into a clean image. \textit{Tree-Ring Watermarking} chooses the initial noise array so that its Fourier transform contains a carefully constructed pattern near its center. This pattern is called the ``key.'' This initial noise vector is then converted into an image using the standard diffusion pipeline with no modifications.  
To detect the watermark in an image, the diffusion model is inverted using the process described in Section \ref{inverse} to retrieve the original noise array used for generation.  This array is then checked to see whether the key is present.

Rather than imprint the key into the Gaussian array directly, which might cause noticeable patterns in the resulting image, we imprint the key into the Fourier transform of the starting noise vector.
We choose a binary mask $M$, and sample the \textit{key} $k^* \in \mathbb{C}^{|M|}$. As such, the initial noise vector $x_T \in \mathbb{R}^L$ can be described in Fourier space as 
\begin{equation}\label{eq:injection}
    \mathcal{F}(x_T)_i \sim
    \begin{cases} k^*_i  \qquad &\textnormal{if} \quad i \in M \\
     \mathcal{N}(0,1) \qquad &\textnormal{otherwise.}
    \end{cases}
\end{equation}
For reasons described below, we choose $M$ as a circular mask with radius $r$ centered on the low-frequency modes. 

\begin{figure}[h!]
    \centering
    \begin{tabular}{ccc}
        No Watermark & Watermarked & 
        Attacked 
        \vspace{1mm}
    \\
        \multicolumn{3}{c}{\textbf{``Anime art of a dog in Shenandoah National Park''}}  \vspace{1mm}
        \\
        \includegraphics[width=0.27\textwidth]{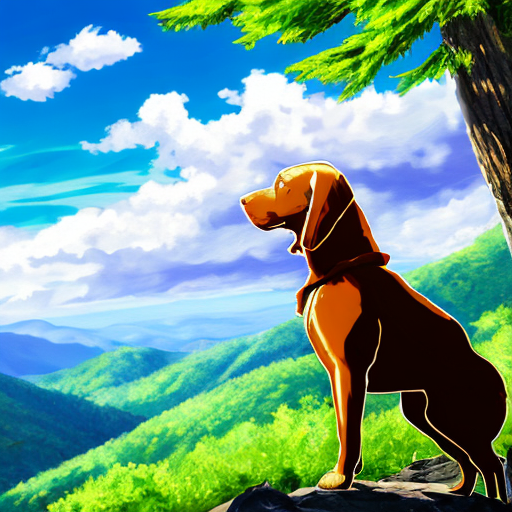}
        & \includegraphics[width=0.27\textwidth]{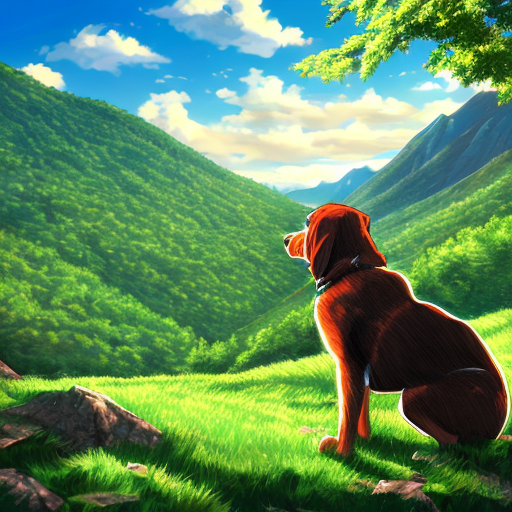}
        & \includegraphics[width=0.27\textwidth]{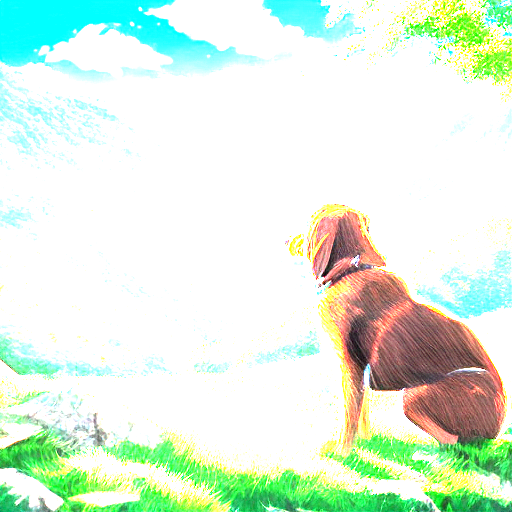}
    \\
        $\text{P-value}=0.27$
        & $3.73\text{e-}60$
        & $7.41\text{e-}16$
        \vspace{1mm}
    \\
        \multicolumn{3}{c}{\textbf{``Synthwave style artwork of a person is kayaking in Acadia National Park''}}  \vspace{1mm}
        \\
        \includegraphics[width=0.27\textwidth]{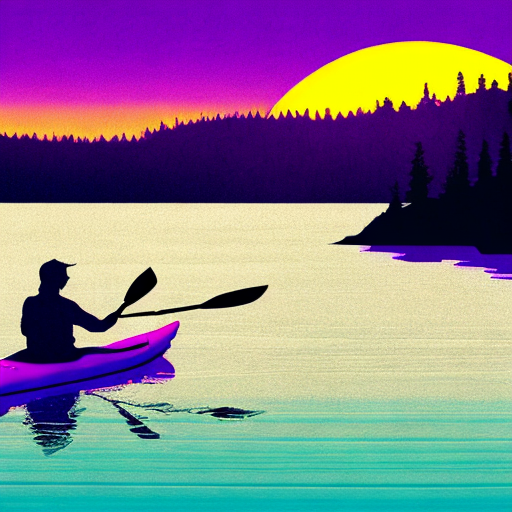}
        & \includegraphics[width=0.27\textwidth]{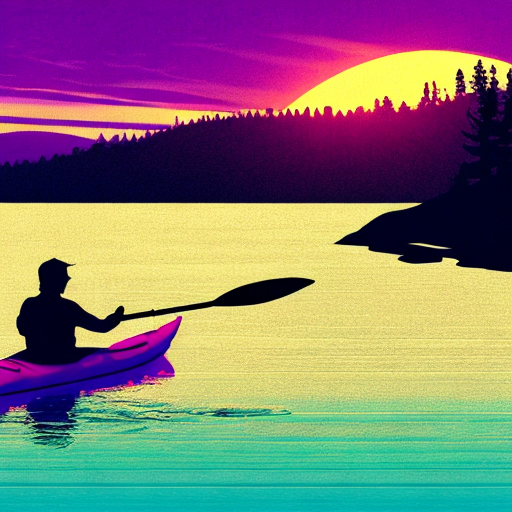}
        & \includegraphics[width=0.27\textwidth]{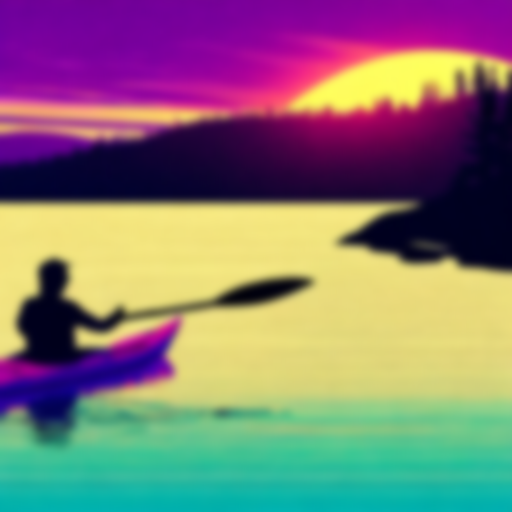}
    \\
        $0.15$
        & $1.38\text{e-}19$
        & $1.51\text{e-}8$
        \vspace{1mm}
    \\
        \multicolumn{3}{c}{\textbf{``An astronaut riding a horse in Zion National Park''}} 
        \\
        \includegraphics[width=0.27\textwidth]{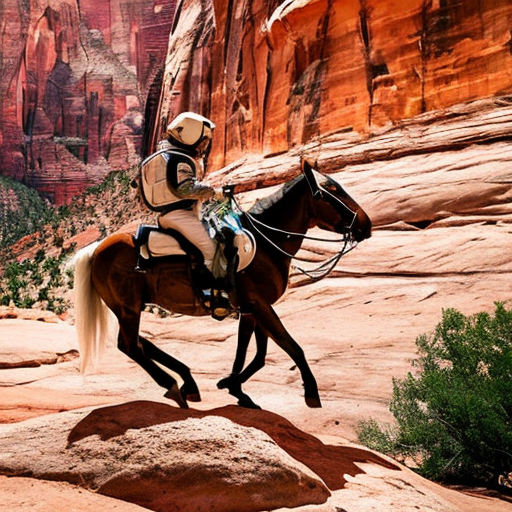}
        & \includegraphics[width=0.27\textwidth]{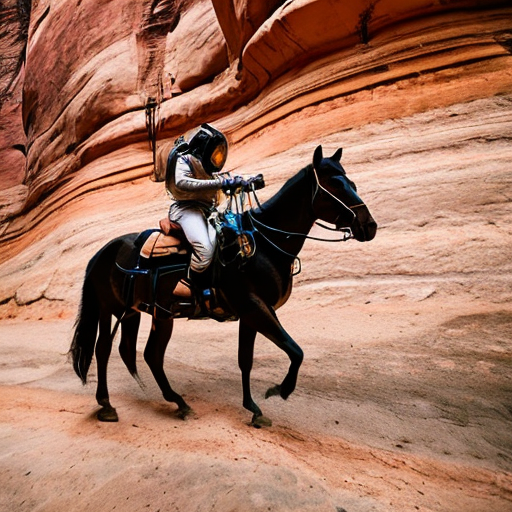}
        & \includegraphics[width=0.27\textwidth]{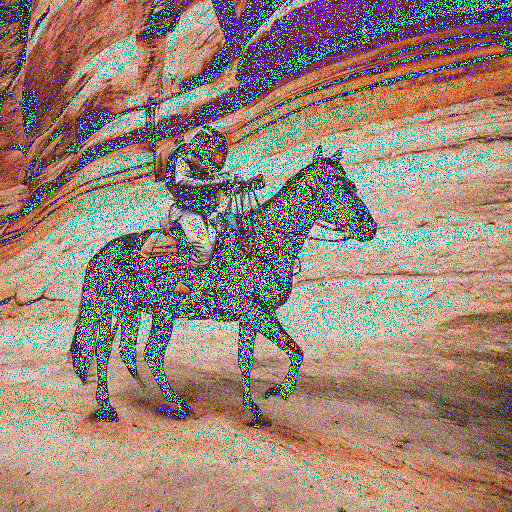}
    \\
        $0.91$
        & $9.91\text{e-}51$
        & $2.90\text{e-}05$
        \vspace{1mm}
        \\
        \multicolumn{3}{c}{\textbf{``A painting of Yosemite National Park in Van Gogh style''}}
        \\
        \includegraphics[width=0.27\textwidth]{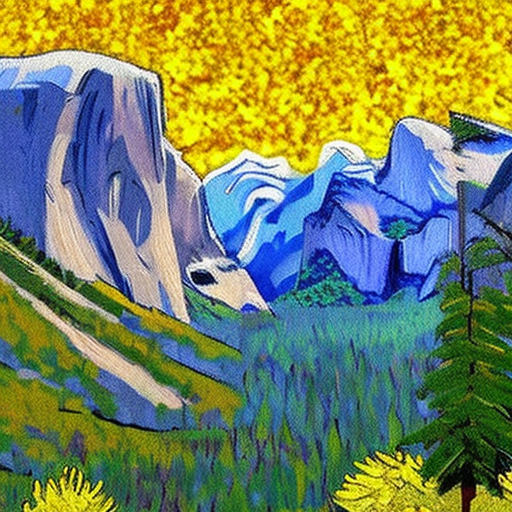}
        & \includegraphics[width=0.27\textwidth]{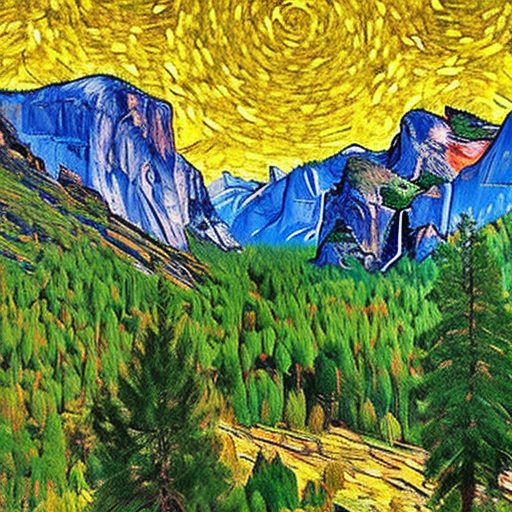}
        & \includegraphics[width=0.27\textwidth]{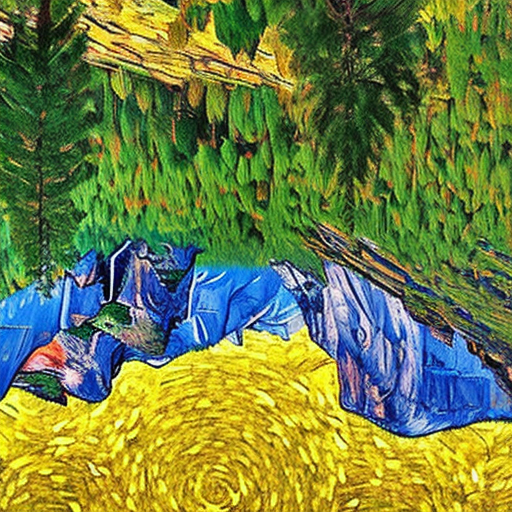} 
    \\
        $0.41$
        & $1.22\text{e-}35$
        & $9.46\text{e-}07$
    \end{tabular}
    \caption{The qualitative results show three types of images: non-watermarked, \textit{Tree-Ring$_{\text{Rings}}$} watermarked, and attacked watermarked images. A P-value is provided below each image, which corresponds to the probability of the detected watermark structure occurring by random chance. From top to bottom, the watermarked images are attacked by color jitter with a brightness factor of $6$, Gaussian blur with an $8\times 8$ filter size, Gaussian noise with $\sigma=0.1$, and a $180^{\circ}$ rotation, respectively.}
    \label{fig:p-value}
\end{figure}

At detection time, given an image $x'_0$, the model owner can obtain an approximated initial noise vector $x'_T$ through the DDIM inversion process:  $x'_T = \mathcal{D}^\dagger_\theta(x'_0)$. The final metric is calculated as the L1 distance between the inverted noise vector and the key in the Fourier space of the watermarked area $M$, i.e. 
\begin{equation}\label{eq:detection-equation}
    d_\text{detection distance} = \frac{1}{|M|} \sum_{i \in M} |k^*_i -   \mathcal{F}(x'_T)_i|,
\end{equation}
and the watermark is detected if this falls below a tuned threshold $\tau$. We later discuss how to calibrate this threshold to a given false-positive either based on a given set of pairs of watermarked and unwatermarked images, or to be set to guarantee a fixed P-value in \cref{sec:pvalue}.

The process described above is straightforward. However, its success depends strongly on the construction of the ``key'' pattern, which we discuss below.

\subsection{Constructing a \textit{Tree-Ring} Key}
We watermark images by placing a ``key'' pattern into the Fourier space of the original Gaussian noise array. 
Our patterns can exploit several classical properties of the Fourier transform for periodic signals that we informally state here.
\begin{itemize}[topsep=1mm, leftmargin=1cm]
\item A rotation in pixel space corresponds to a rotation in Fourier space.
\item A translation in pixel space multiplies all Fourier coefficients by a constant complex number.
\item A dilation/compression in pixel space corresponds to a compression/dilation in Fourier space.
\item Color jitter in pixel space (adding a constant to all pixels in a channel) corresponds to changing the magnitude of the zero-frequency Fourier mode.
\end{itemize}
A number of classical watermarking strategies rely on watermarking in Fourier space and exploit similar invariances \citep{pitas_method_1998,solachidis_circularly_2001}.  Our watermark departs from classical methods by applying a Fourier watermark to a random noise array {\em before}  diffusion takes place.  Curiously, we will observe below that the invariant properties above are preserved in $x_T$ even when image manipulations are done in pixel space of $x_0$. 

In addition to exploiting the invariances above, the chosen key should also be statistically similar to Gaussian noise.  Note that the Fourier transform of a Gaussian noise array is also distributed as Gaussian noise. For this reason, choosing a highly non-Gaussian key may cause a distribution shift that impacts the diffusion model. 

We consider three different types of keys, with the respective benefits of each pattern being demonstrated in subsequent experimental sections. We believe there are numerous other interesting and practical types that can be explored in future work.

\textbf{\textit{Tree-Ring$_{\text{Zeros}}$}:} We choose the mask to be a circular region to preserve invariance to rotations in image space.  The key is chosen to be an array of zeros, which creates invariance to shifts, crops, and dilations. This key is invariant to manipulations, but at the cost of departing severely from the Gaussian distribution. It also prevents multiple keys from being used to distinguish between models.

\textbf{\textit{Tree-Ring$_{\text{Rand}}$}:} We draw the a fixed key $k^*$ from a Gaussian distribution. The key has the same iid Gaussian nature as the original Fourier modes of the noise array, and so we anticipate this strategy will have the least impact on generation quality.
This method also offers the flexibility for the model owner to possess multiple keys.  However, it is not invariant to make image manipulations.

\textbf{\textit{Tree-Ring$_{\text{Rings}}$}:} We introduce a pattern comprised of \textit{multiple rings}, and constant value along each ring. This makes the watermark invariant to rotations.
We choose the constant ring values from a Gaussian distribution. This provides some invariance to multiple types of image transforms, while also ensuring that the overall distribution is only minimally shifted from an isotropic Gaussian.

\subsection{Deriving P-values for Watermark Detection}\label{sec:pvalue}

\begin{figure}
    \centering
    \vspace{-0.5cm}
    \includegraphics[width=0.5\textwidth]{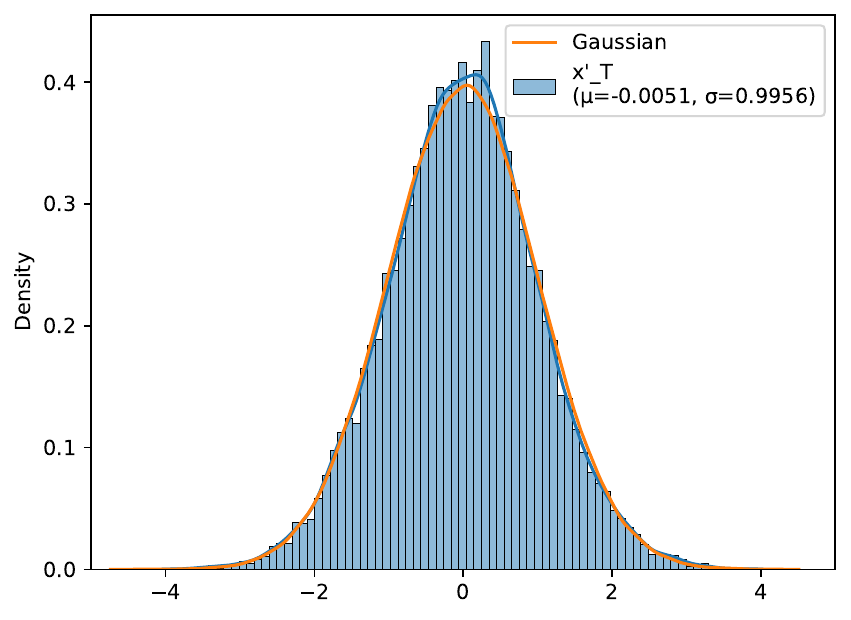}
    \caption{Histogram of the array $x_T'$ obtained for a natural image, which is Gaussian.}
    \label{fig:noise-hist}
\end{figure}

A key desideratum for a reliable watermark detector is that it provide an interpretable P-value that communicates to the user how likely it is that the observed watermark could have occurred in a natural image by random chance. In addition to making detection results interpretable, P-values can be used to set the threshold of detection, i.e., the watermark is ``detected'' when $p$ is below a chosen threshold $\alpha$. By doing so, one can explicitly control the false positive rate $\alpha,$ making false accusations statistically unlikely.  

To this end, we construct a statistical test for the presence of the watermark that produces a rigorous P-value. The forward diffusion process is designed to map images onto Gaussian noise, and so we assume a null hypothesis in which the entries in the array $x_T'$ obtained for a natural image are Gaussian. We find that this assumption holds quite well in practice, see \cref{fig:noise-hist}.

For any test image $x_0'$, we compute the approximate initial vector $x_T'$ and then set $y= \mathcal{F}(x'_T)$. We then define the following null hypothesis
\begin{equation}
    \operatorname{H}_0: \textit{$y$ is drawn from a Gaussian distribution $\mathcal{N}(\mathbf{0}, \sigma^2 I_\mathbb{C})$.}
\end{equation}
Here,  $\sigma^2$ is an unknown variance, which we estimate for each image\footnote{Our statistical test is only sensitive to \textit{Tree-Ring$_{\text{Rand}}$} and \textit{Tree-Ring$_{\text{Rings}}$}.   \textit{Tree-Ring$_{\text{Zeros}}$} results in the pathological case that $\sigma \approx 0$ for watermarked images resulting in overly conservative/large P-values.} using the formula $\sigma^2 = \frac{1}{M} \sum_{i \in M} |y_i|^2.$
To test this hypothesis, we define the score
\begin{equation}
    \eta = \frac{1}{\sigma^2} \sum_{i \in M} |k_i^* -  y |^2. 
\end{equation}
Our formula for $\eta$ is closely related to \cref{eq:detection-equation}, but we switch to a sum-of-squares metric and remove the variance from $y$ to simplify statistical analysis. When $H_0$ is true, the distribution of $\eta$ is exactly a {\em noncentral} $\chi^2$ {\em distribution} \citep{patnaik_non-central_1949}, with $|M|$ degrees of freedom and non-centrality parameter $\lambda=\frac{1}{\sigma^2}\sum_{i} |k_i^*|^2$. 

We declare an image to be watermarked if the value of $\eta$ is too small to occur by random chance. 
The probability of observing a value as small as $\eta$ is given by the cumulative distribution function $\Phi_{\chi^2}$  of the noncentral $\chi^2$ distribution:
\begin{equation}
    p=\Pr \left(\chi^2_{|M|, \lambda}\leq \eta \middle| H_{0} \right) = \Phi_{\chi^2}(z).
\end{equation}
$\Phi_{\chi^2}$ is a standard statistical function 
\citep{glasserman_monte_2003}, 
available in \texttt{scipy} and many other statistics libraries. 
%


We show qualitative examples of the proposed watermarking scheme and accompanying P-values in \cref{fig:p-value}.  For each prompt, we show the generated image with and without the watermark, and also a watermarked image subjected to a transformation.  For each image, we report a P-value.  As expected, these values are large for non-watermarked images, and small (enabling rejection of the null hypothesis) when the watermark is present.  Transformations reduce the watermark strength as reflected in the increased P-value.

\section{Experiments}
We perform experiments on two common diffusion models to measure the efficacy and reliability of the \textit{Tree-Ring Watermarking} technique across diverse attack scenarios. Furthermore, we carry out ablation studies to provide an in-depth exploration of this technique.

\subsection{Experimental Setting}
We employ Stable Diffusion-v2 \citep{rombach_high-resolution_2022}, an open-source, state-of-the-art latent text-to-image diffusion model, along with a $256\times256$ ImageNet diffusion model\footnote{\url{https://github.com/openai/guided-diffusion}} \citep{dhariwal_diffusion_2021}. In the main experiment, we use $50$ inference steps for generation and detection for both models. For Stable Diffusion, we use the default guidance scale of $7.5$, and we use an empty prompt for DDIM inversion, emulating that the image prompt would be unknown at detection time. The watermark radius $r$ we use is $10$. Later, we conduct more ablation studies on these important hyperparameters. All experiments are conducted on a single NVIDIA RTX A4000.

Our comparative analysis includes three baselines: two training-free methods, DwtDct and DwtDctSvd \citep{10.5555/1564551}, and a pre-trained GAN-based watermarking model, RivaGAN \citep{zhang_robust_2019, NIPS2014_5ca3e9b1}. However, these baseline methods are designed for steganography, which conceals a target bit-string within an image. To ensure a fair comparison with our exclusively watermarking method, we employ the distance between the decoded bit-string and the target bit-string (Bit Accuracy) as the measurement metric. The approach of \citet{10.5555/1564551} is currently deployed as a watermark mechanism in Stable Diffusion\footnote{\url{github.com/CompVis/stable-diffusion/blob/main/scripts/txt2img.py\#L69}}.

\subsection{Benchmarking Watermark Accuracy and Image Quality}
To benchmark the effectiveness of the watermark, we primarily report the area under the curve (\textbf{AUC}) of the receiver operating characteristic (ROC) curve, and the True Positive Rate when the False Positive Rate is at $1\%$, denoted as $\bf \textbf{TPR}\boldsymbol{@1\%}\textbf{FPR}$. To demonstrate the generation quality of the watermarked images, we assess the Frechet Inception Distance (\textbf{FID}) \citep{Heusel2017GANsTB} for both models. Additionally, for the Stable Diffusion model, we also evaluate the \textbf{CLIP score} \citep{Radford2021LearningTV} between the generated image and the prompt, as measured by OpenCLIP-ViT/G \citep{Cherti2022ReproducibleSL}. For AUC and $\text{TPR}@1\%\text{FPR}$, we create $1,000$ watermarked and $1,000$ unwatermarked images for each run. For FID, we generate $5,000$ images for Stable Diffusion and $10,000$ images for the ImageNet Model. The FID of Stable Diffusion is evaluated on the MS-COCO-2017 training dataset \citep{Lin2014MicrosoftCC}, and the FID of the ImageNet Model is gauged on the ImageNet-1k training dataset \citep{5206848}. All reported metrics are averaged across $5$ runs using different random seeds following this protocol.

In \cref{table:main_table}, we present the main experimental results for Stable Diffusion and the ImageNet model. In the clean setting, all baselines except DwtDct and all \textit{Tree-Ring Watermarking} variants are strongly detectable. \textit{Tree-Ring$_{\text{Rand}}$} and \textit{Tree-Ring$_{\text{Rings}}$} show negligible impact on the FID and no impact on the CLIP score.

\begin{table}
\centering
\renewcommand{\arraystretch}{1.3}
\caption{Main Results. $\text{T}@1\%\text{F}$ represents $\text{TPR}@1\%\text{FPR}$. We evaluate watermark accuracy in both benign and adversarial settings. Adversarial here refers to average performance over a battery of image manipulations. An extended version with additional details and standard error estimates can be found in Supplementary Material.}
\label{table:main_table}
\begin{tabular}{cccccc}
\toprule
Model & Method & \begin{tabular}[c]{@{}c@{}}AUC/\text{T}@1\%\text{F}\\ (Clean)\end{tabular} & \begin{tabular}[c]{@{}c@{}}AUC/$\text{T}@1\%\text{F}$\\ (Adversarial)\end{tabular} & FID $\downarrow$ & CLIP Score $\uparrow$ \\ \hline
\multirow{6}{*}{\begin{tabular}[c]{@{}c@{}}\textbf{Stable Diff.}\\ $\text{FID}=25.29$\\ $\text{CLIP Score}$\\$=0.363$\end{tabular}} & DwtDct & $0.974$ / $0.624$ & $0.574$ / $0.092$ & $25.10_{.09}$ & $0.362_{.000}$ \\
 & DwtDctSvd & $1.000$ / $1.000$ & $0.702$ / $0.262$ & $25.01_{.09}$ & $0.359_{.000}$ \\
 & RivaGAN & $0.999$ / $0.999$ & $0.854$ / $0.448$ & $\mathbf{24.51_{.17}}$ & $0.361_{.000}$ \\ \cline{2-6}
 & \textbf{\textit{Tree-Ring$_{\text{Zeros}}$}} & $0.999$ / $0.999$ & $0.963$ / $\mathbf{0.715}$ & $26.56_{.07}$ & $0.356_{.000}$ \\
 & \textbf{\textit{Tree-Ring$_{\text{Rand}}$}} & $1.000$ / $1.000$ & $0.918$ / $0.702$ & $25.47_{.05}$ & $0.363_{.001}$ \\
 & \textbf{\textit{Tree-Ring$_{\text{Rings}}$}} & $1.000$ / $1.000$ & $\mathbf{0.975}$ / $0.694$ & $25.93_{.13}$ & $\mathbf{0.364_{.000}}$ \\ \midrule
\multirow{6}{*}{\begin{tabular}[c]{@{}c@{}}\textbf{ImageNet}\\ $\text{FID}=17.73$\end{tabular}} & DwtDct & $0.899$ / $0.244$ & $0.536$ / $0.037$ & $17.77_{.01}$ & - \\
 & DwtDctSvd & $1.000$ / $1.000$ & $0.713$ / $0.187$ & $18.55_{.02}$ & - \\
 & RivaGAN & $1.000$ / $1.000$ & $0.882$ / $0.509$ & $18.70_{.02}$ & - \\ \cline{2-6}
 & \textbf{\textit{Tree-Ring$_{\text{Zeros}}$}} & $0.999$ / $1.000$ & $0.921$ / $0.476$ & $18.78_{.00}$ & - \\
 & \textbf{\textit{Tree-Ring$_{\text{Rand}}$}} & $0.999$ / $1.000$ & $0.940$ / $0.585$ & $18.68_{.09}$ & - \\
 & \textbf{\textit{Tree-Ring$_{\text{Rings}}$}} & $0.999$ / $0.999$ & $\mathbf{0.966}$ / $\mathbf{0.603}$ & $\mathbf{17.68_{.16}}$ & -\\ \bottomrule
\end{tabular}
\vspace{-.5cm}
\end{table}

\subsection{Benchmarking Watermark Robustness}
To benchmark the robustness of our watermark, we focus on documenting its performance under $6$ prevalent data augmentations utilized as attacks. These include $75^{\circ}$ rotation, $25\%$ JPEG compression, $75\%$ random cropping and scaling, Gaussian blur with an $8\times 8$ filter size, Gaussian noise with $\sigma=0.1$, and color jitter with a brightness factor uniformly sampled between $0$ and $6$. Additionally, we conduct ablation studies to investigate the impact of varying intensities of these attacks.
We report both AUC and $\text{TPR}@1\%\text{FPR}$ in the average case where we average the metrics over the clean setting and all attacks. In all ablation studies, we report the average case.

In \cref{table:main_table}, the baseline methods fail in the presence of adversaries. On the contrary, our methods demonstrate higher reliability in adversarial settings. Among them, \textit{Tree-Ring${_\text{Rings}}$} performs the best under adversarial conditions, a result of our careful watermark pattern design.

Further, we show the AUC for each attack setting in \cref{table:main_attack}. Notably, \textit{Tree-Ring$_{\text{Zeros}}$} demonstrates high robustness against most perturbations, except for Gaussian noise and color jitter. Similarly, \textit{Tree-Ring$_{\text{Rand}}$} is robust in most scenarios but performs poorly when faced with rotation, as expected. Overall, \textit{Tree-Ring$_{\text{Rings}}$} delivers the best average performance while offering the model owner the flexibility of multiple different random keys. It is worth noting that the baseline method RivaGan also demonstrates strong robustness in most scenarios, but it is important to highlight that our method is training-free and really ``invisible''.

\begin{table}
\centering
\caption{AUC under each Attack for Stable Diffusion, showing the effectiveness of \textit{Tree-Ring$_{\text{Rings}}$} over a number of augmentations. Cr. \& Sc. refers to random cropping and rescaling. Additional results for the ImageNet model can be found in Supplementary Material.}
\label{table:main_attack}
\begin{tabular}{ccccccccc}
\toprule
Method & Clean & Rotation & JPEG & Cr. \& Sc. & Blurring & Noise & Color Jitter & Avg \\ \midrule
DwtDct & $0.974$ & $0.596$& $0.492$ & $0.640$ & $0.503$ & $0.293$ & $0.519$ & $0.574$ \\
DwtDctSvd & $\mathbf{1.000}$ & $0.431$ & $0.753$ & $0.511$ & $0.979$ & $0.706$ & $0.517$ & $0.702$ \\
RivaGan & $0.999$ & $0.173$ & $0.981$ & $\mathbf{0.999}$ & $0.974$ & $0.888$ & $0.963$ & $0.854$ \\ \midrule
\textbf{\textit{T-R$_{\text{Zeros}}$}} & $0.999$ & $\mathbf{0.994}$ & $0.984$ & $\mathbf{0.999}$ & $0.977$ & $0.877$ & $0.907$ & $0.963$ \\
\textbf{\textit{T-R$_{\text{Rand}}$}} & $\mathbf{1.000}$ & $0.486$ & $\mathbf{0.999}$ & $0.971$ & $\mathbf{0.999}$ & $\mathbf{0.972}$ & $\mathbf{0.994}$ & $0.918$ \\
\textbf{\textit{T-R$_{\text{Rings}}$}} & $\mathbf{1.000}$ & $0.935$ & $\mathbf{0.999}$ & $0.961$ & $\mathbf{0.999}$ & $0.944$ & $0.983$ & $\mathbf{0.975}$ \\
\bottomrule
\end{tabular}
\vspace{-.25cm}
\end{table}

\subsection{Ablation Experiments}
In this section, we undertake exhaustive ablation studies with the Ring pattern on several key hyperparameters to demonstrate the efficacy of \textit{Tree-Ring Watermarking}. Except for the ablation on attacks, the reported numbers represent averages over all attack scenarios and clean images.

\begin{wrapfigure}[18]{r}{0.6\textwidth}
    \centering
    \vspace{-0.75cm}
    \includegraphics[width=0.6\textwidth]{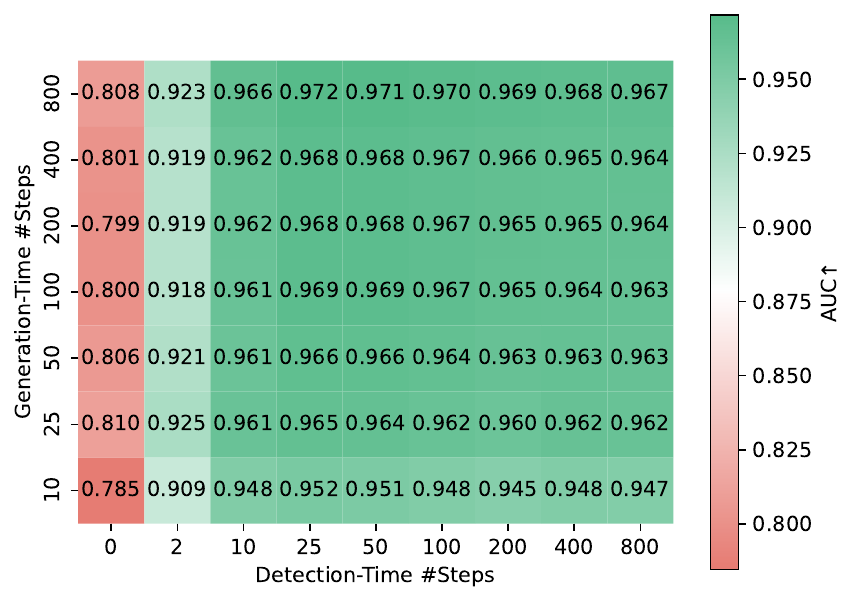}
    \vspace{-0.5cm}
    \caption{Ablation on Number of Generation Steps versus Detection Steps. Detection succeeds independent of the number of DDIM used to generate data.}
    \label{fig:abl-steps}
    \vspace{-0.5cm}
\end{wrapfigure} 
In \cref{fig:abl-steps}, we compare AUC across all step combinations. Surprisingly, even with a significant difference between the generation-time and detection-time $\#$steps, the decrease in AUC is minimal when the model owner uses a reasonable number of inference steps for detection without knowledge of the true generation-time steps. This indicates that the DDIM inversion maintains its robustness in approximating the initial noise vector, and is effective for watermark detection irrespective of the exact number of steps employed. Interestingly, we notice a trend where the detection power appears to be slightly stronger with fewer inference steps at detection time or a larger number of inference steps at generation time. This is an advantageous scenario as the model owner now does not actually need to carry out a large number of steps for DDIM inversion, while concurrently, the model owner (or the user) is free to choose the number of generation steps that achieve the best quality \citep{rombach_high-resolution_2022}.

\textbf{Number of Steps Used for Generation and Detection.} A key unknown variable for the model owner at the detection time is the actual number of inference steps used during the generation time. This factor could potentially impact the precision of the DDIM inversion approximation of the initial noise vector. To scrutinize this, we systematically vary the number of steps for both the generation and detection time.
Due to the computational demands of sampling with a high number of inference steps, we employ a total of $400$ images for each run.

\begin{figure}
    \centering
    \subfigure[Ablation on Watermark Radii]{\includegraphics[height=0.3\textwidth]{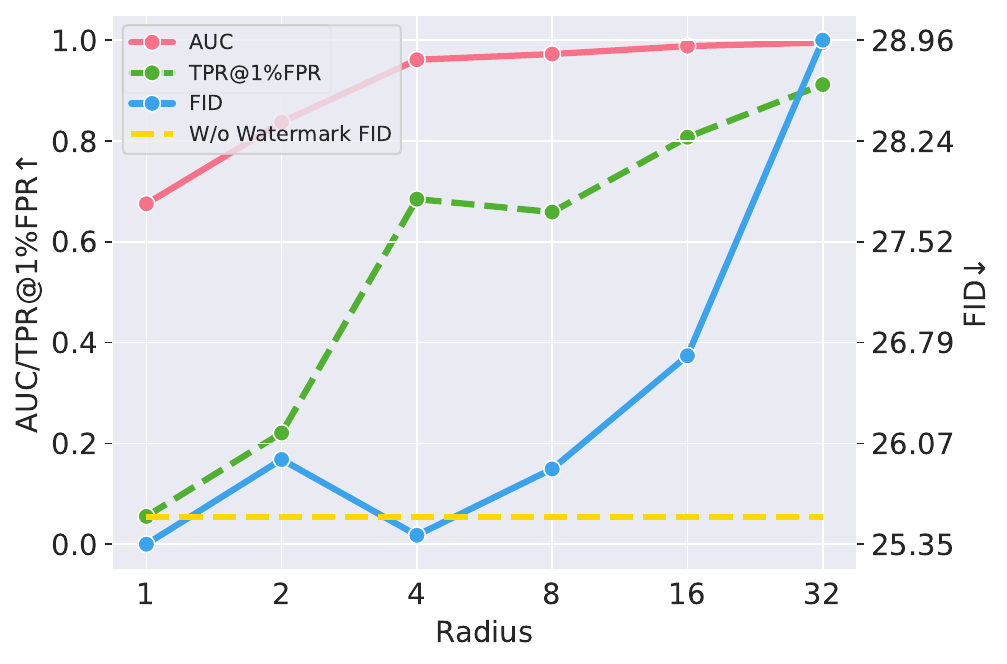}\label{fig:abl-radii}}
    \subfigure[Ablation on Guidance Scales]{\includegraphics[height=0.3\textwidth]{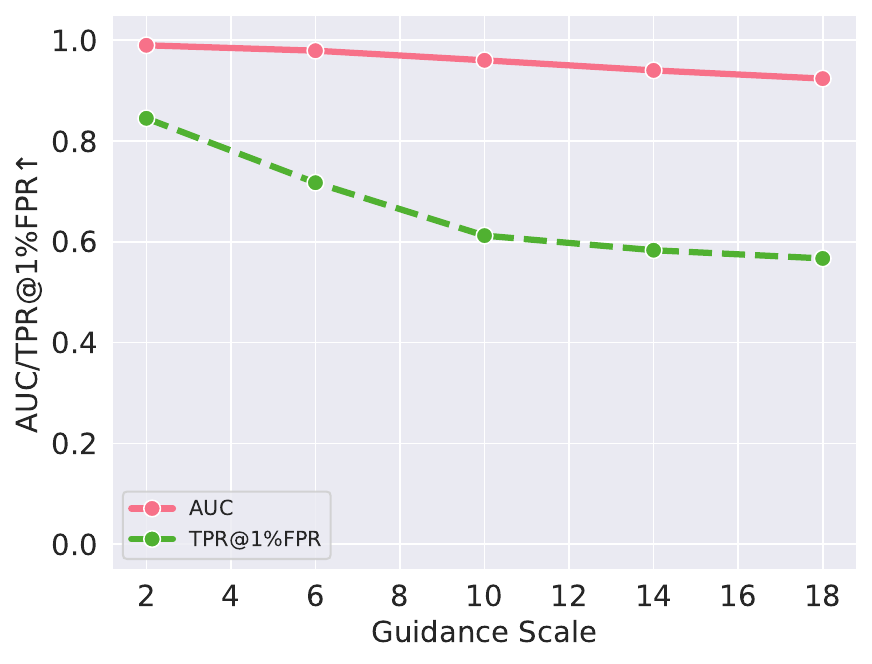}\label{fig:abl-guidance}}
    \caption{Ablation on Watermark Radii and Guidance Scales.}
    \label{fig:abl-radii-guidance}
    \vspace{-.5cm}
\end{figure}

\textbf{Watermark radii.} The radius of injected watermarking patterns is another critical hyperparameter affecting robustness and generation quality. The corresponding results are shown in Figure \ref{fig:abl-radii}. As the watermarking radius increases, the watermark's robustness improves. Nevertheless, there is a trade-off with generation quality. We overall confirm a radius of $16$ to provide reasonably low FID while maintaining strong detection power.

\textbf{Guidance scales.} Guidance scale is a hyperparameter that controls the significance of the text condition. Higher guidance scales mean the generation more strictly adheres to the text guidance, whereas lower guidance scales provide the model with greater creative freedom. Optimal guidance scales typically range between $5$ and $15$ for the Stable Diffusion model we employ. We explore this factor from $2$ to $18$ in Figure \ref{fig:abl-guidance} and highlight that the strength of the guidance is always unknown during detection time. Although a higher guidance scale does increase the error for DDIM inversion due to the lack of this ground-truth guidance during detection, the watermark remains robust and reliable even at a guidance scale of $18$. This is again beneficial for practical purposes, allowing the model owner to keep guidance scale a tunable setting for their users.

\begin{figure}
    \centering
    \subfigure[Rotation]{\includegraphics[height=0.25\textwidth]{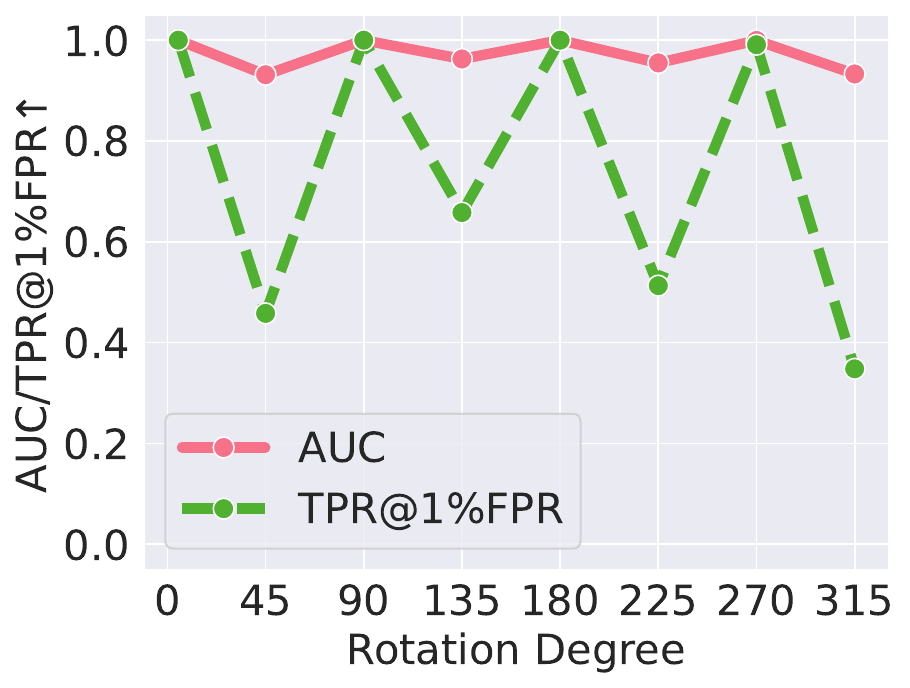}\label{fig:abl-rotate}}
    \subfigure[JPEG Compression]{\includegraphics[height=0.25\textwidth]{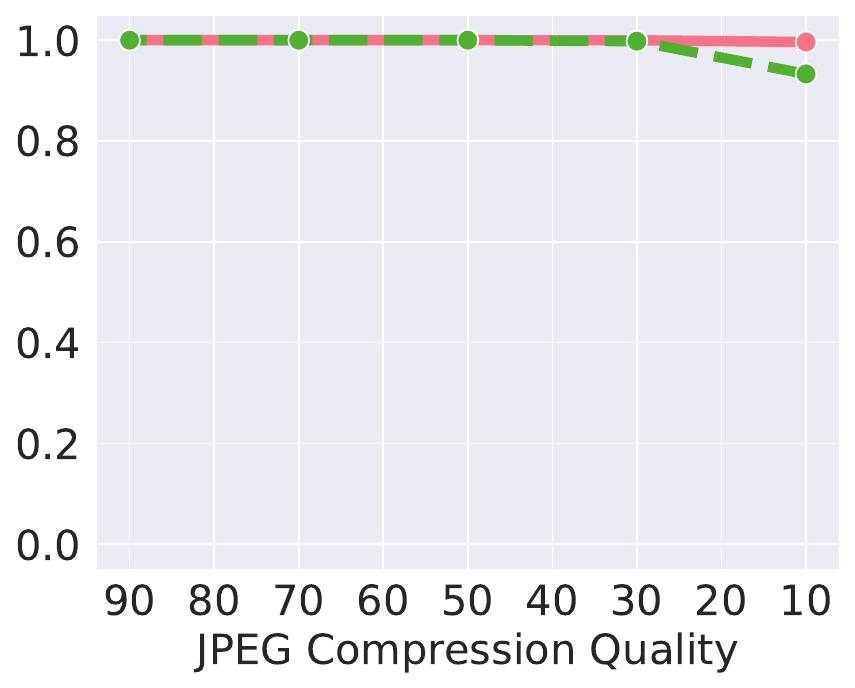}\label{fig:abl-jpeg}}
    \subfigure[Cropping + Scaling]{\includegraphics[height=0.25\textwidth]{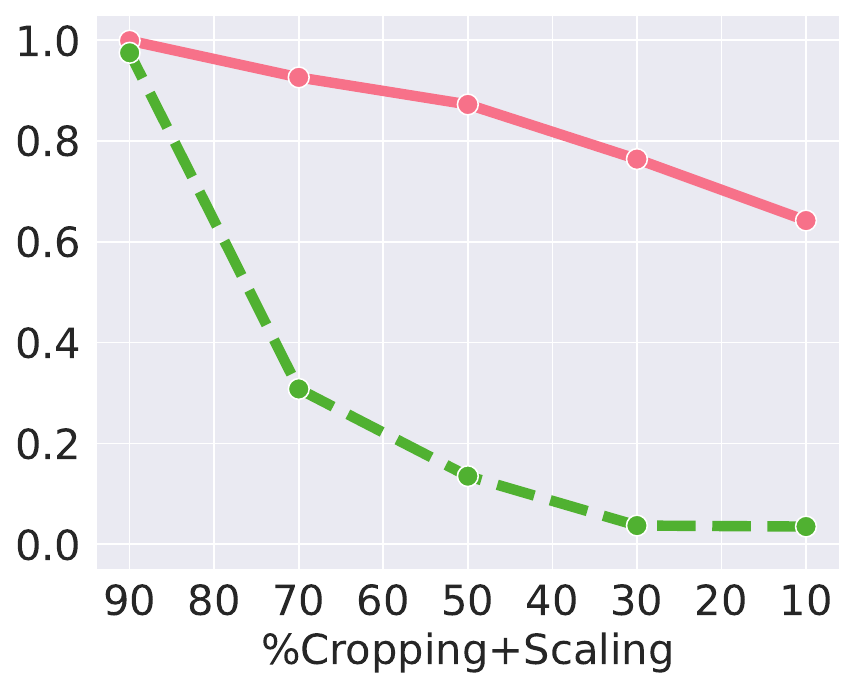}\label{fig:abl-crop}}
    \subfigure[Gaussian Blurring]{\includegraphics[height=0.25\textwidth]{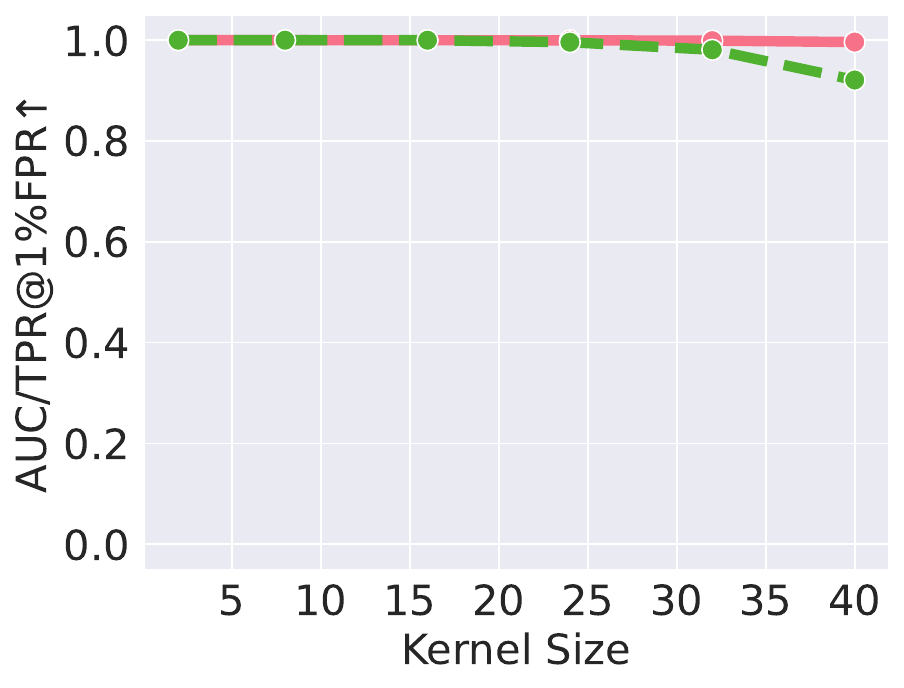}\label{fig:abl-blur}}
    \subfigure[Gaussian Noise]{\includegraphics[height=0.25\textwidth]{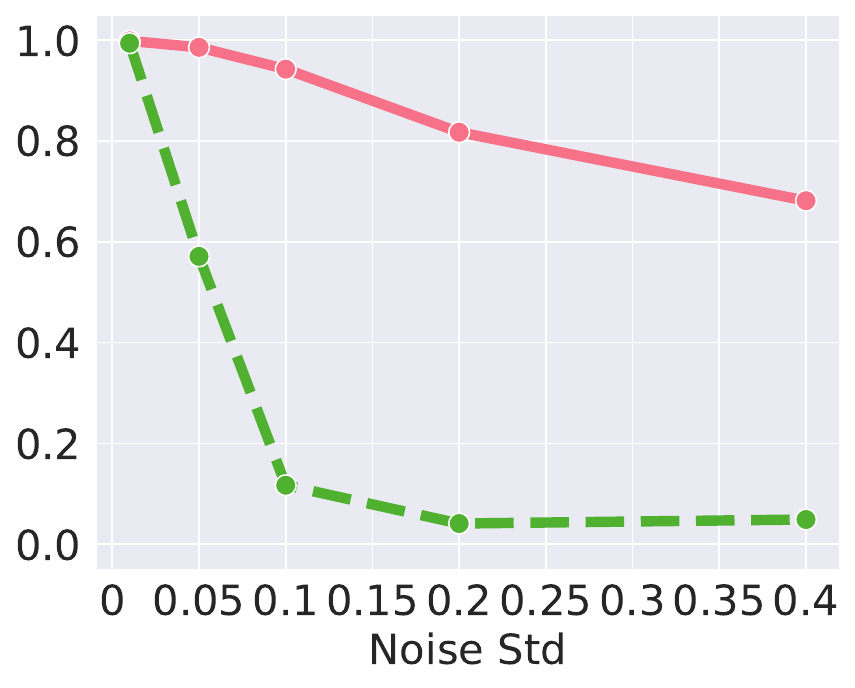}\label{fig:abl-noise}}
    \subfigure[Color Jitter]{\includegraphics[height=0.25\textwidth]{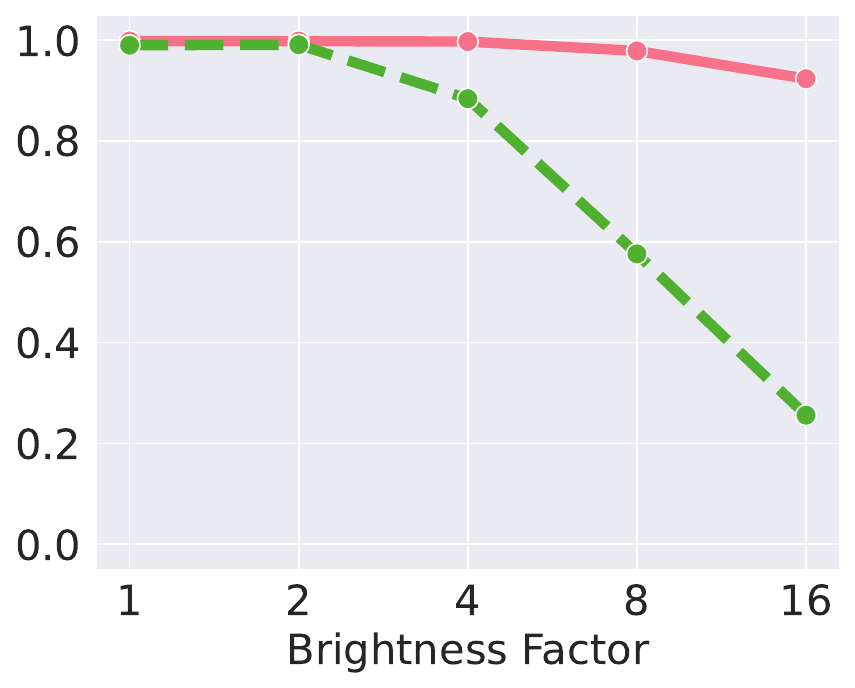}\label{fig:abl-color}}
    \caption{Ablation on Different Perturbation Strengths.}
    \label{fig:abl-attack-power}
    \vspace{-.5cm}
\end{figure}

\textbf{Attack strengths.} Further, we test out the robustness of \textit{Tree-Ring Watermarking} under each attack with various attack strengths. As shown in \cref{fig:abl-attack-power}, even with extreme perturbations like Gaussian blurring with kernel size $40$, \textit{Tree-Ring Watermarking} can still be reliably detected. 

\section{Limitations and Future Work}
\textit{Tree-Ring Watermarking} requires the model owner to use DDIM during inference. Today, DDIM is still likely the most popular sampling method due to its economical use of GPU resources and high quality. However, the proposed watermark will need to be adapted to other sampling schemes should DDIM fall out of favor.
Further, the proposed watermark is by design only verifiable by the model owner because model parameters are needed to perform the inversion process. This has advantages against adversaries, who cannot perform a white-box attack on the watermark or even verify whether an ensemble of manipulations broke the watermark. However, it also restricts third parties from detecting the watermark without relying on an API. Finally, it is currently not yet clear how large the capacity for multiple keys $k^*$ would be. Would it be possible to assign a unique key to every user of the API?

The effectiveness of the proposed watermark is directly related to the accuracy of the inverse DDIM process. Future work that improves the accuracy of this inversion \citep{zhang_robustness_2023}, or utilizes invertible diffusion models as described in \citet{wallace_edict_2022}, would also improve watermarking power further.

\section{Conclusion}
We propose a new approach to watermarking generative diffusion models using minimal shifts of their output distribution. This leads to watermarks that are truly invisible on a per-sample basis. We describe how to optimally shift, so that the watermark remains detectable even under strong image manipulations that might be encountered in daily usage and handling of generated images.

\section{Acknowledgements}
This work was made possible by the ONR MURI program, DARPA GARD (HR00112020007), the Office of Naval Research (N000142112557), and the AFOSR MURI program.  Commercial support was provided by Capital One Bank, the Amazon Research Award program, and Open Philanthropy. Further support was provided by the National Science Foundation (IIS-2212182), and by the NSF TRAILS Institute (2229885).

\bibliography{manual_references,automatic_references}

\begin{thebibliography}{43}
\providecommand{\natexlab}[1]{#1}
\providecommand{\url}[1]{\texttt{#1}}
\expandafter\ifx\csname urlstyle\endcsname\relax
  \providecommand{\doi}[1]{doi: #1}\else
  \providecommand{\doi}{doi: \begingroup \urlstyle{rm}\Url}\fi

\bibitem[{Al-Haj}(2007)]{al-haj_combined_2007}
Ali {Al-Haj}.
\newblock Combined {{DWT-DCT Digital Image Watermarking}}.
\newblock \emph{Journal of Computer Science}, 3\penalty0 (9):\penalty0
  740--746, September 2007.
\newblock ISSN 15493636.
\newblock \doi{10.3844/jcssp.2007.740.746}.
\newblock URL \url{http://www.thescipub.com/abstract/?doi=jcssp.2007.740.746}.

\bibitem[Bansal et~al.(2022)Bansal, Chiang, Curry, Jain, Wigington, Manjunatha,
  Dickerson, and Goldstein]{bansal_certified_2022}
Arpit Bansal, Ping-Yeh Chiang, Michael~J. Curry, Rajiv Jain, Curtis Wigington,
  Varun Manjunatha, John~P. Dickerson, and Tom Goldstein.
\newblock Certified {{Neural Network Watermarks}} with {{Randomized
  Smoothing}}.
\newblock In \emph{Proceedings of the 39th {{International Conference}} on
  {{Machine Learning}}}, pages 1450--1465. {PMLR}, June 2022.
\newblock URL \url{https://proceedings.mlr.press/v162/bansal22a.html}.

\bibitem[Bender et~al.(2021)Bender, Gebru, {McMillan-Major}, and
  Shmitchell]{bender_dangers_2021}
Emily~M. Bender, Timnit Gebru, Angelina {McMillan-Major}, and Shmargaret
  Shmitchell.
\newblock On the {{Dangers}} of {{Stochastic Parrots}}: {{Can Language Models
  Be Too Big}}?
\newblock In \emph{Proceedings of the 2021 {{ACM Conference}} on {{Fairness}},
  {{Accountability}}, and {{Transparency}}}, {{FAccT}} '21, pages 610--623,
  {New York, NY, USA}, March 2021. {Association for Computing Machinery}.
\newblock ISBN 978-1-4503-8309-7.
\newblock \doi{10.1145/3442188.3445922}.
\newblock URL \url{https://doi.org/10.1145/3442188.3445922}.

\bibitem[Boland(1996)]{boland_watermarking_1996}
Francis~Morgan Boland.
\newblock Watermarking digital images for copyright protection.
\newblock 1996.
\newblock URL \url{http://www.tara.tcd.ie/handle/2262/19682}.

\bibitem[Chang et~al.(2005)Chang, Tsai, and Lin]{chang_svd-based_2005}
Chin-Chen Chang, Piyu Tsai, and Chia-Chen Lin.
\newblock {{SVD-based}} digital image watermarking scheme.
\newblock \emph{Pattern Recognition Letters}, 26\penalty0 (10):\penalty0
  1577--1586, July 2005.
\newblock ISSN 0167-8655.
\newblock \doi{10.1016/j.patrec.2005.01.004}.
\newblock URL
  \url{https://www.sciencedirect.com/science/article/pii/S0167865505000140}.

\bibitem[Cherti et~al.(2022)Cherti, Beaumont, Wightman, Wortsman, Ilharco,
  Gordon, Schuhmann, Schmidt, and Jitsev]{Cherti2022ReproducibleSL}
Mehdi Cherti, Romain Beaumont, Ross Wightman, Mitchell Wortsman, Gabriel
  Ilharco, Cade Gordon, Christoph Schuhmann, Ludwig Schmidt, and Jenia Jitsev.
\newblock Reproducible scaling laws for contrastive language-image learning.
\newblock \emph{ArXiv}, abs/2212.07143, 2022.

\bibitem[Cox et~al.(1996)Cox, Kilian, Leighton, and Shamoon]{cox_secure_1996}
I.J. Cox, J.~Kilian, T.~Leighton, and T.~Shamoon.
\newblock Secure spread spectrum watermarking for images, audio and video.
\newblock \emph{Proceedings of 3rd IEEE International Conference on Image
  Processing}, 3:\penalty0 243--246, 1996.
\newblock \doi{10.1109/ICIP.1996.560429}.
\newblock URL \url{http://ieeexplore.ieee.org/document/560429/}.

\bibitem[Cox et~al.(2007)Cox, Miller, Bloom, Fridrich, and
  Kalker]{10.5555/1564551}
Ingemar Cox, Matthew Miller, Jeffrey Bloom, Jessica Fridrich, and Ton Kalker.
\newblock \emph{Digital Watermarking and Steganography}.
\newblock Morgan Kaufmann Publishers Inc., San Francisco, CA, USA, 2 edition,
  2007.
\newblock ISBN 9780080555805.

\bibitem[Deng et~al.(2009)Deng, Dong, Socher, Li, Li, and Fei-Fei]{5206848}
Jia Deng, Wei Dong, Richard Socher, Li-Jia Li, Kai Li, and Li~Fei-Fei.
\newblock Imagenet: A large-scale hierarchical image database.
\newblock In \emph{2009 IEEE Conference on Computer Vision and Pattern
  Recognition}, pages 248--255, 2009.
\newblock \doi{10.1109/CVPR.2009.5206848}.

\bibitem[Dhariwal and Nichol(2021)]{dhariwal_diffusion_2021}
Prafulla Dhariwal and Alex Nichol.
\newblock Diffusion {{Models Beat GANs}} on {{Image Synthesis}}.
\newblock \emph{arxiv:2105.05233[cs, stat]}, June 2021.
\newblock \doi{10.48550/arXiv.2105.05233}.
\newblock URL \url{http://arxiv.org/abs/2105.05233}.

\bibitem[Fei et~al.(2022)Fei, Xia, Tondi, and Barni]{fei_supervised_2022}
Jianwei Fei, Zhihua Xia, Benedetta Tondi, and Mauro Barni.
\newblock Supervised {{GAN Watermarking}} for {{Intellectual Property
  Protection}}.
\newblock \emph{arxiv:2209.03466[cs]}, September 2022.
\newblock \doi{10.48550/arXiv.2209.03466}.
\newblock URL \url{http://arxiv.org/abs/2209.03466}.

\bibitem[Fernandez et~al.(2023)Fernandez, Couairon, J{\'e}gou, Douze, and
  Furon]{fernandez_stable_2023}
Pierre Fernandez, Guillaume Couairon, Herv{\'e} J{\'e}gou, Matthijs Douze, and
  Teddy Furon.
\newblock The {{Stable Signature}}: {{Rooting Watermarks}} in {{Latent
  Diffusion Models}}.
\newblock \emph{arxiv:2303.15435[cs]}, March 2023.
\newblock \doi{10.48550/arXiv.2303.15435}.
\newblock URL \url{http://arxiv.org/abs/2303.15435}.

\bibitem[Glasserman(2003)]{glasserman_monte_2003}
Paul Glasserman.
\newblock \emph{Monte {{Carlo Methods}} in {{Financial Engineering}}},
  volume~53 of \emph{Stochastic {{Modelling}} and {{Applied Probability}}}.
\newblock {Springer}, {New York, NY}, 2003.
\newblock ISBN 978-1-4419-1822-2 978-0-387-21617-1.
\newblock \doi{10.1007/978-0-387-21617-1}.
\newblock URL \url{http://link.springer.com/10.1007/978-0-387-21617-1}.

\bibitem[Goodfellow et~al.(2014)Goodfellow, Pouget-Abadie, Mirza, Xu,
  Warde-Farley, Ozair, Courville, and Bengio]{NIPS2014_5ca3e9b1}
Ian Goodfellow, Jean Pouget-Abadie, Mehdi Mirza, Bing Xu, David Warde-Farley,
  Sherjil Ozair, Aaron Courville, and Yoshua Bengio.
\newblock Generative adversarial nets.
\newblock In Z.~Ghahramani, M.~Welling, C.~Cortes, N.~Lawrence, and K.Q.
  Weinberger, editors, \emph{Advances in Neural Information Processing
  Systems}, volume~27. Curran Associates, Inc., 2014.
\newblock URL
  \url{https://proceedings.neurips.cc/paper_files/paper/2014/file/5ca3e9b122f61f8f06494c97b1afccf3-Paper.pdf}.

\bibitem[Grinbaum and Adomaitis(2022)]{grinbaum_ethical_2022}
Alexei Grinbaum and Laurynas Adomaitis.
\newblock The {{Ethical Need}} for {{Watermarks}} in {{Machine-Generated
  Language}}.
\newblock \emph{arxiv:2209.03118[cs]}, September 2022.
\newblock \doi{10.48550/arXiv.2209.03118}.
\newblock URL \url{http://arxiv.org/abs/2209.03118}.

\bibitem[Hayes and Danezis(2017)]{hayes_generating_2017}
Jamie Hayes and George Danezis.
\newblock Generating steganographic images via adversarial training.
\newblock In \emph{Advances in {{Neural Information Processing Systems}}},
  volume~30. {Curran Associates, Inc.}, 2017.
\newblock URL
  \url{https://papers.nips.cc/paper\_files/paper/2017/hash/fe2d010308a6b3799a3d9c728ee74244-Abstract.html}.

\bibitem[Heusel et~al.(2017)Heusel, Ramsauer, Unterthiner, Nessler, and
  Hochreiter]{Heusel2017GANsTB}
Martin Heusel, Hubert Ramsauer, Thomas Unterthiner, Bernhard Nessler, and Sepp
  Hochreiter.
\newblock Gans trained by a two time-scale update rule converge to a local nash
  equilibrium.
\newblock In \emph{NIPS}, 2017.

\bibitem[Ho et~al.(2020)Ho, Jain, and Abbeel]{ho_denoising_2020}
Jonathan Ho, Ajay Jain, and Pieter Abbeel.
\newblock Denoising {{Diffusion Probabilistic Models}}.
\newblock In \emph{Advances in {{Neural Information Processing Systems}}},
  volume~33, pages 6840--6851. {Curran Associates, Inc.}, 2020.
\newblock URL
  \url{https://proceedings.neurips.cc/paper/2020/hash/4c5bcfec8584af0d967f1ab10179ca4b-Abstract.html}.

\bibitem[Kirchenbauer et~al.(2023)Kirchenbauer, Geiping, Wen, Katz, Miers, and
  Goldstein]{kirchenbauer_watermark_2023}
John Kirchenbauer, Jonas Geiping, Yuxin Wen, Jonathan Katz, Ian Miers, and Tom
  Goldstein.
\newblock A {{Watermark}} for {{Large Language Models}}.
\newblock \emph{arxiv:2301.10226[cs]}, January 2023.
\newblock \doi{10.48550/arXiv.2301.10226}.
\newblock URL \url{http://arxiv.org/abs/2301.10226}.

\bibitem[Kutter and Petitcolas(1999)]{kutter_fair_1999}
Martin Kutter and Fabien A.~P. Petitcolas.
\newblock Fair benchmark for image watermarking systems.
\newblock In \emph{Security and {{Watermarking}} of {{Multimedia Contents}}},
  volume 3657, pages 226--239. {SPIE}, April 1999.
\newblock \doi{10.1117/12.344672}.
\newblock URL
  \url{https://www.spiedigitallibrary.org/conference-proceedings-of-spie/3657/0000/Fair-benchmark-for-image-watermarking-systems/10.1117/12.344672.full}.

\bibitem[Langelaar et~al.(2000)Langelaar, Setyawan, and
  Lagendijk]{langelaar_watermarking_2000}
G.C. Langelaar, I.~Setyawan, and R.L. Lagendijk.
\newblock Watermarking digital image and video data. {{A}} state-of-the-art
  overview.
\newblock \emph{IEEE Signal Processing Magazine}, 17\penalty0 (5):\penalty0
  20--46, September 2000.
\newblock ISSN 1558-0792.
\newblock \doi{10.1109/79.879337}.

\bibitem[Lin et~al.(2014)Lin, Maire, Belongie, Hays, Perona, Ramanan,
  Doll{\'a}r, and Zitnick]{Lin2014MicrosoftCC}
Tsung-Yi Lin, Michael Maire, Serge~J. Belongie, James Hays, Pietro Perona, Deva
  Ramanan, Piotr Doll{\'a}r, and C.~Lawrence Zitnick.
\newblock Microsoft coco: Common objects in context.
\newblock In \emph{European Conference on Computer Vision}, 2014.

\bibitem[Liu et~al.(2019)Liu, Huang, Luo, Cao, Yang, Wei, and
  Zhou]{liu_optimized_2019}
Junxiu Liu, Jiadong Huang, Yuling Luo, Lvchen Cao, Su~Yang, Duqu Wei, and
  Ronglong Zhou.
\newblock An {{Optimized Image Watermarking Method Based}} on {{HD}} and
  {{SVD}} in {{DWT Domain}}.
\newblock \emph{IEEE Access}, 7:\penalty0 80849--80860, 2019.
\newblock ISSN 2169-3536.
\newblock \doi{10.1109/ACCESS.2019.2915596}.

\bibitem[Nichol and Dhariwal(2021)]{nichol_improved_2021}
Alex Nichol and Prafulla Dhariwal.
\newblock Improved {{Denoising Diffusion Probabilistic Models}}.
\newblock \emph{arxiv:2102.09672[cs, stat]}, February 2021.
\newblock \doi{10.48550/arXiv.2102.09672}.
\newblock URL \url{http://arxiv.org/abs/2102.09672}.

\bibitem[O'Ruanaidh and Pun(1997)]{oruanaidh_rotation_1997}
J.J.K. O'Ruanaidh and T.~Pun.
\newblock Rotation, scale and translation invariant digital image watermarking.
\newblock In \emph{Proceedings of {{International Conference}} on {{Image
  Processing}}}, volume~1, pages 536--539 vol.1, October 1997.
\newblock \doi{10.1109/ICIP.1997.647968}.

\bibitem[Patnaik(1949)]{patnaik_non-central_1949}
P.~B. Patnaik.
\newblock The {{Non-Central}} {$X$}2- and {{F-Distribution}} and their
  {{Applications}}.
\newblock \emph{Biometrika}, 36\penalty0 (1/2):\penalty0 202--232, 1949.
\newblock ISSN 0006-3444.
\newblock \doi{10.2307/2332542}.
\newblock URL \url{https://www.jstor.org/stable/2332542}.

\bibitem[Pitas(1998)]{pitas_method_1998}
I.~Pitas.
\newblock A method for watermark casting on digital image.
\newblock \emph{IEEE Transactions on Circuits and Systems for Video
  Technology}, 8\penalty0 (6):\penalty0 775--780, October 1998.
\newblock ISSN 1558-2205.
\newblock \doi{10.1109/76.728421}.

\bibitem[Radford et~al.(2021)Radford, Kim, Hallacy, Ramesh, Goh, Agarwal,
  Sastry, Askell, Mishkin, Clark, Krueger, and
  Sutskever]{Radford2021LearningTV}
Alec Radford, Jong~Wook Kim, Chris Hallacy, Aditya Ramesh, Gabriel Goh,
  Sandhini Agarwal, Girish Sastry, Amanda Askell, Pamela Mishkin, Jack Clark,
  Gretchen Krueger, and Ilya Sutskever.
\newblock Learning transferable visual models from natural language
  supervision.
\newblock In \emph{International Conference on Machine Learning}, 2021.

\bibitem[Rombach et~al.(2022)Rombach, Blattmann, Lorenz, Esser, and
  Ommer]{rombach_high-resolution_2022}
Robin Rombach, Andreas Blattmann, Dominik Lorenz, Patrick Esser, and Bj{\"o}rn
  Ommer.
\newblock High-{{Resolution Image Synthesis With Latent Diffusion Models}}.
\newblock In \emph{Proceedings of the {{IEEE}}/{{CVF Conference}} on {{Computer
  Vision}} and {{Pattern Recognition}}}, pages 10684--10695, 2022.
\newblock URL
  \url{https://openaccess.thecvf.com/content/CVPR2022/html/Rombach\_High-Resolution\_Image\_Synthesis\_With\_Latent\_Diffusion\_Models\_CVPR\_2022\_paper.html}.

\bibitem[Seo et~al.(2004)Seo, Haitsma, Kalker, and Yoo]{seo_robust_2004}
Jin~S Seo, Jaap Haitsma, Ton Kalker, and Chang~D Yoo.
\newblock A robust image fingerprinting system using the {{Radon}} transform.
\newblock \emph{Signal Processing: Image Communication}, 19\penalty0
  (4):\penalty0 325--339, April 2004.
\newblock ISSN 0923-5965.
\newblock \doi{10.1016/j.image.2003.12.001}.
\newblock URL
  \url{https://www.sciencedirect.com/science/article/pii/S0923596503001541}.

\bibitem[Solachidis and Pitas(2001)]{solachidis_circularly_2001}
V.~Solachidis and L.~Pitas.
\newblock Circularly symmetric watermark embedding in 2-{{D DFT}} domain.
\newblock \emph{IEEE Transactions on Image Processing}, 10\penalty0
  (11):\penalty0 1741--1753, November 2001.
\newblock ISSN 1941-0042.
\newblock \doi{10.1109/83.967401}.

\bibitem[Song and Ermon(2019)]{song_generative_2019}
Yang Song and Stefano Ermon.
\newblock Generative {{Modeling}} by {{Estimating Gradients}} of the {{Data
  Distribution}}.
\newblock \emph{arXiv:1907.05600 [cs, stat]}, October 2019.
\newblock URL \url{http://arxiv.org/abs/1907.05600}.

\bibitem[Song and Ermon(2020)]{song_improved_2020}
Yang Song and Stefano Ermon.
\newblock Improved {{Techniques}} for {{Training Score-Based Generative
  Models}}.
\newblock \emph{arXiv:2006.09011 [cs, stat]}, June 2020.
\newblock URL \url{http://arxiv.org/abs/2006.09011}.

\bibitem[Uchida et~al.(2017)Uchida, Nagai, Sakazawa, and
  Satoh]{uchida_embedding_2017}
Yusuke Uchida, Yuki Nagai, Shigeyuki Sakazawa, and Shin'ichi Satoh.
\newblock Embedding {{Watermarks}} into {{Deep Neural Networks}}.
\newblock In \emph{Proceedings of the 2017 {{ACM}} on {{International
  Conference}} on {{Multimedia Retrieval}}}, pages 269--277, {Bucharest
  Romania}, June 2017. {ACM}.
\newblock ISBN 978-1-4503-4701-3.
\newblock \doi{10.1145/3078971.3078974}.
\newblock URL \url{https://dl.acm.org/doi/10.1145/3078971.3078974}.

\bibitem[Wallace et~al.(2022)Wallace, Gokul, and Naik]{wallace_edict_2022}
Bram Wallace, Akash Gokul, and Nikhil Naik.
\newblock {{EDICT}}: {{Exact Diffusion Inversion}} via {{Coupled
  Transformations}}.
\newblock \emph{arxiv:2211.12446[cs]}, December 2022.
\newblock \doi{10.48550/arXiv.2211.12446}.
\newblock URL \url{http://arxiv.org/abs/2211.12446}.

\bibitem[Wan et~al.(2022)Wan, Wang, Zhang, Li, Yu, and
  Sun]{wan_comprehensive_2022}
Wenbo Wan, Jun Wang, Yunming Zhang, Jing Li, Hui Yu, and Jiande Sun.
\newblock A comprehensive survey on robust image watermarking.
\newblock \emph{Neurocomputing}, 488:\penalty0 226--247, June 2022.
\newblock ISSN 0925-2312.
\newblock \doi{10.1016/j.neucom.2022.02.083}.
\newblock URL
  \url{https://www.sciencedirect.com/science/article/pii/S0925231222002533}.

\bibitem[Yu et~al.(2022)Yu, Skripniuk, Abdelnabi, and
  Fritz]{yu_artificial_2022}
Ning Yu, Vladislav Skripniuk, Sahar Abdelnabi, and Mario Fritz.
\newblock Artificial {{Fingerprinting}} for {{Generative Models}}: {{Rooting
  Deepfake Attribution}} in {{Training Data}}.
\newblock \emph{arxiv:2007.08457[cs]}, March 2022.
\newblock \doi{10.48550/arXiv.2007.08457}.
\newblock URL \url{http://arxiv.org/abs/2007.08457}.

\bibitem[Zeng et~al.(2023)Zeng, Zhou, Xue, and Patel]{zeng2023securing}
Yu~Zeng, Mo~Zhou, Yuan Xue, and Vishal~M Patel.
\newblock Securing deep generative models with universal adversarial signature.
\newblock \emph{arXiv preprint arXiv:2305.16310}, 2023.

\bibitem[Zhang et~al.(2018)Zhang, Gu, Jang, Wu, Stoecklin, Huang, and
  Molloy]{zhang_protecting_2018}
Jialong Zhang, Zhongshu Gu, Jiyong Jang, Hui Wu, Marc Stoecklin, Heqing Huang,
  and Ian Molloy.
\newblock Protecting intellectual property of deep neural networks with
  watermarking.
\newblock In \emph{{{ACM Symposium}} on {{Information}}, {{Computer}} and
  {{Communications Security}}}. {Association for Computing Machinery, Inc.},
  May 2018.
\newblock ISBN 978-1-4503-5576-6.
\newblock \doi{10.1145/3196494.3196550}.
\newblock URL
  \url{https://research.ibm.com/publications/protecting-intellectual-property-of-deep-neural-networks-with-watermarking}.

\bibitem[Zhang et~al.(2023)Zhang, Das, and Kumar]{zhang_robustness_2023}
Jiaxin Zhang, Kamalika Das, and Sricharan Kumar.
\newblock On the {{Robustness}} of {{Diffusion Inversion}} in {{Image
  Manipulation}}.
\newblock In \emph{{{ICLR}} 2023 {{Workshop}} on {{Trustworthy}} and {{Reliable
  Large-Scale Machine Learning Models}}}, April 2023.
\newblock URL \url{https://openreview.net/forum?id=fr8kurMWJIP}.

\bibitem[Zhang et~al.(2019)Zhang, Xu, {Cuesta-Infante}, and
  Veeramachaneni]{zhang_robust_2019}
Kevin~Alex Zhang, Lei Xu, Alfredo {Cuesta-Infante}, and Kalyan Veeramachaneni.
\newblock Robust {{Invisible Video Watermarking}} with {{Attention}}.
\newblock \emph{arxiv:1909.01285[cs]}, September 2019.
\newblock \doi{10.48550/arXiv.1909.01285}.
\newblock URL \url{http://arxiv.org/abs/1909.01285}.

\bibitem[Zhao et~al.(2023)Zhao, Pang, Du, Yang, Cheung, and
  Lin]{zhao_recipe_2023}
Yunqing Zhao, Tianyu Pang, Chao Du, Xiao Yang, Ngai-Man Cheung, and Min Lin.
\newblock A {{Recipe}} for {{Watermarking Diffusion Models}}.
\newblock \emph{arxiv:2303.10137[cs]}, March 2023.
\newblock \doi{10.48550/arXiv.2303.10137}.
\newblock URL \url{http://arxiv.org/abs/2303.10137}.

\bibitem[Zhu et~al.(2018)Zhu, Kaplan, Johnson, and {Fei-Fei}]{zhu_hidden_2018}
Jiren Zhu, Russell Kaplan, Justin Johnson, and Li~{Fei-Fei}.
\newblock {{HiDDeN}}: {{Hiding Data}} with {{Deep Networks}}.
\newblock In \emph{Proceedings of the {{European Conference}} on {{Computer
  Vision}} ({{ECCV}})}, pages 657--672, 2018.
\newblock URL
  \url{https://openaccess.thecvf.com/content\_ECCV\_2018/html/Jiren\_Zhu\_HiDDeN\_Hiding\_Data\_ECCV\_2018\_paper.html}.

\end{thebibliography}
\bibliographystyle{plainnat}

\newpage

\appendix
\section{Appendix} \label{app:appendix}

\begin{table}[h]
\centering
\renewcommand{\arraystretch}{1.3}
\caption{Main Results with Error Bars. $\text{T}@1\%\text{F}$ represents $\text{TPR}@1\%\text{FPR}$. We evaluate watermark accuracy in both benign and adversarial settings. Adversarial here refers to average performance over a battery of image manipulations.}
\label{table:main_table_error_bar}
\begin{tabular}{cccccc}
\toprule
Model & Method & \begin{tabular}[c]{@{}c@{}}AUC/\text{T}@1\%\text{F}\\ (Clean)\end{tabular} & \begin{tabular}[c]{@{}c@{}}AUC/$\text{T}@1\%\text{F}$\\ (Adversarial)\end{tabular} & FID $\downarrow$ & CLIP Score $\uparrow$ \\ \hline
\multirow{6}{*}{\begin{tabular}[c]{@{}c@{}}\textbf{Stable Diff.}\\ $\text{FID}=25.29$\\ $\text{CLIP Score}$\\$=0.363$\end{tabular}} & DwtDct & $0.974_{.001}$ / $0.624_{.013}$ & $0.574_{.005}$ / $0.092_{.004}$ & $25.10_{.09}$ & $0.362_{.000}$ \\
 & DwtDctSvd & $1.000_{.000}$ / $1.000_{.000}$ & $0.702_{.000}$ / $0.262_{.011}$ & $25.01_{.09}$ & $0.359_{.000}$ \\
 & RivaGAN & $0.999_{.000}$ / $0.999_{.000}$ & $0.854_{.002}$ / $0.448_{.006}$ & $\mathbf{24.51_{.17}}$ & $0.361_{.000}$ \\ \cline{2-6}
 & \textbf{\textit{T-R$_{\text{Zeros}}$}} & $0.999_{.000}$ / $0.999_{.000}$ & $0.963_{.001}$ / $\mathbf{0.715_{.021}}$ & $26.56_{.07}$ & $0.356_{.000}$ \\
 & \textbf{\textit{T-R$_{\text{Rand}}$}} & $1.000_{.000}$ / $1.000_{.000}$ & $0.918_{.005}$ / $0.702_{.017}$ & $25.47_{.05}$ & $0.363_{.001}$ \\
 & \textbf{\textit{T-R$_{\text{Rings}}$}} & $1.000_{.000}$ / $1.000_{.000}$ & $\mathbf{0.975_{.001}}$ / $0.694_{.018}$ & $25.93_{.13}$ & $\mathbf{0.364_{.000}}$ \\ \midrule
\multirow{6}{*}{\begin{tabular}[c]{@{}c@{}}\textbf{ImageNet}\\ $\text{FID}=17.73$\end{tabular}} & DwtDct & $0.899_{.040}$ / $0.244_{.203}$ & $0.536_{.016}$ / $0.037_{.029}$ & $17.77_{.01}$ & - \\
 & DwtDctSvd & $1.000_{.000}$ / $1.000_{.000}$ & $0.713_{.019}$ / $0.187_{.008}$ & $18.55_{.02}$ & - \\
 & RivaGAN & $1.000_{.000}$ / $1.000_{.000}$ & $0.882_{.010}$ / $0.509_{.009}$ & $18.70_{.02}$ & - \\ \cline{2-6}
 & \textbf{\textit{T-R$_{\text{Zeros}}$}} & $0.999_{.000}$ / $1.000_{.000}$ & $0.921_{.000}$ / $0.476_{.000}$ & $18.78_{.00}$ & - \\
 & \textbf{\textit{T-R$_{\text{Rand}}$}} & $0.999_{.000}$ / $1.000_{.000}$ & $0.940_{.004}$ / $0.585_{.006}$ & $18.68_{.09}$ & - \\
 & \textbf{\textit{T-R$_{\text{Rings}}$}} & $0.999_{.000}$ / $0.999_{.000}$ & $\mathbf{0.966_{.005}}$ / $\mathbf{0.603_{.006}}$ & $\mathbf{17.68_{.16}}$ & -\\ \bottomrule
\end{tabular}
\end{table}

\begin{table}[h]
\centering
\caption{AUC under each Attack for the ImageNet model, showing the effectiveness of \textit{Tree-Ring$_{\text{Rings}}$} over a number of augmentations. Cr. \& Sc. refers to random cropping and rescaling.}
\label{table:main_attack_imagenet}
\begin{tabular}{ccccccccc}
\toprule
Method & Clean & Rotation & JPEG & Cr. \& Sc. & Blurring & Noise & Color Jitter & Avg \\ \midrule
DwtDct & $0.899$ & $0.478$ & $0.522$ & $0.433$ & $0.512$ & $0.365$ & $0.538$ & $0.536$ \\
DwtDctSvd & $\mathbf{1.000}$ & $0.669$ & $0.568$ & $0.614$ & $0.947$ & $0.656$ & $0.535$ & $0.713$ \\
RivaGan & $\mathbf{1.000}$ & $0.321$ & $\mathbf{0.978}$ & $\mathbf{0.999}$ & $0.988$ & $0.962$ & $0.924$ & $0.882$ \\ \midrule
\textbf{\textit{T-R$_{\text{Zeros}}$}} & $0.999$ & $0.953$ & $0.806$ & $0.997$ & $\mathbf{0.999}$ & $0.938$ & $0.775$ & $0.921$ \\
\textbf{\textit{T-R$_{\text{Rand}}$}} & $0.999$ & $0.682$ & $0.962$ & $0.997$ & $\mathbf{0.999}$ & $\mathbf{0.986}$ & $\mathbf{0.956}$ & $0.940$ \\
\textbf{\textit{T-R$_{\text{Rings}}$}} & $0.999$ & $\mathbf{0.975}$ & $0.940$ & $0.994$ & $\mathbf{0.999}$ & $0.979$ & $0.861$ & $\mathbf{0.966}$ \\
\bottomrule
\end{tabular}
\end{table}

\begin{figure*}[!ht]
    \centering
    \begin{tabular}{cccc}
        W/o Watermark & \textit{Tree-Ring$_{\text{Zeros}}$} & \textit{Tree-Ring$_{\text{Rand}}$} & \textit{Tree-Ring$_{\text{Rings}}$}
        \\
        \includegraphics[scale=0.16]{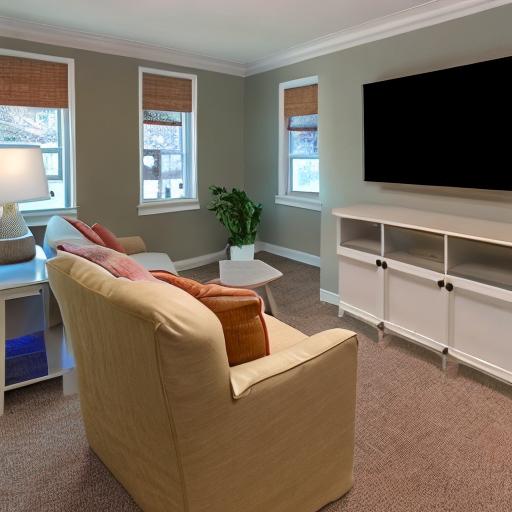} & 
        \includegraphics[scale=0.16]{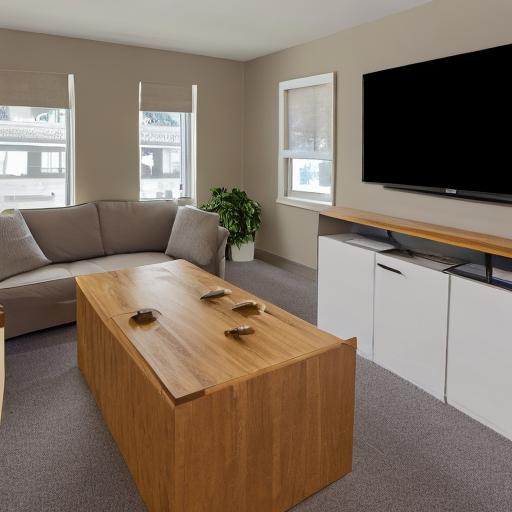} &
        \includegraphics[scale=0.16]{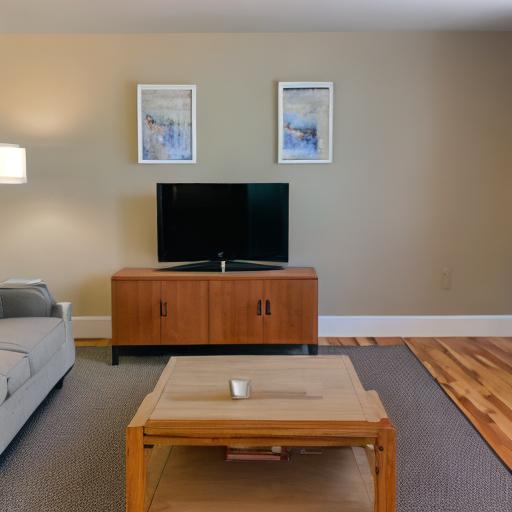} &
        \includegraphics[scale=0.16]{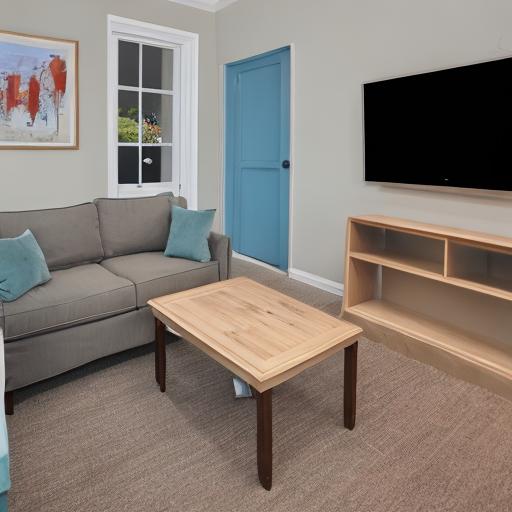}
        \\
        \includegraphics[scale=0.16]{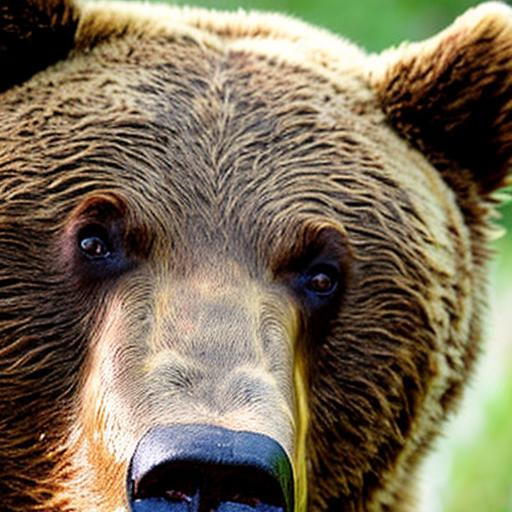} & 
        \includegraphics[scale=0.16]{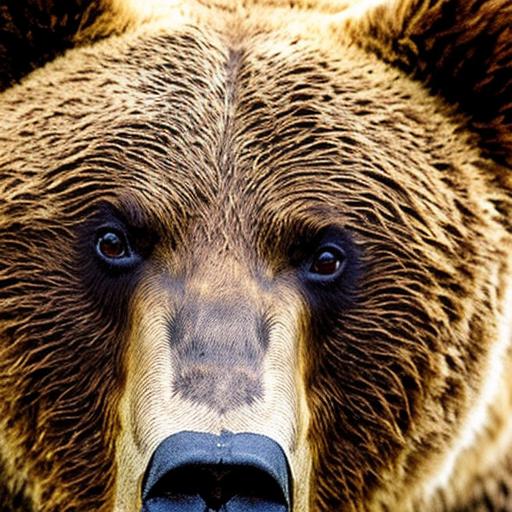} &
        \includegraphics[scale=0.16]{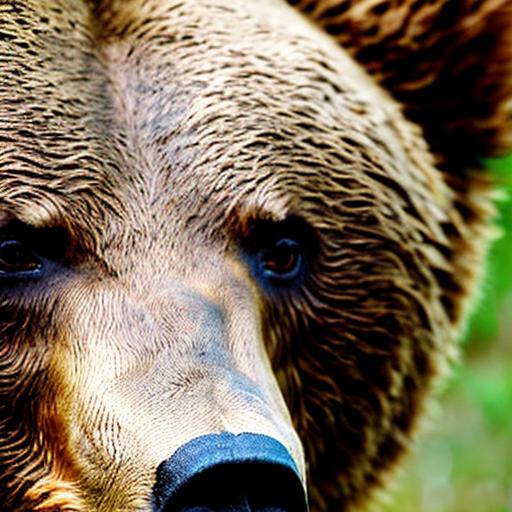} &
        \includegraphics[scale=0.16]{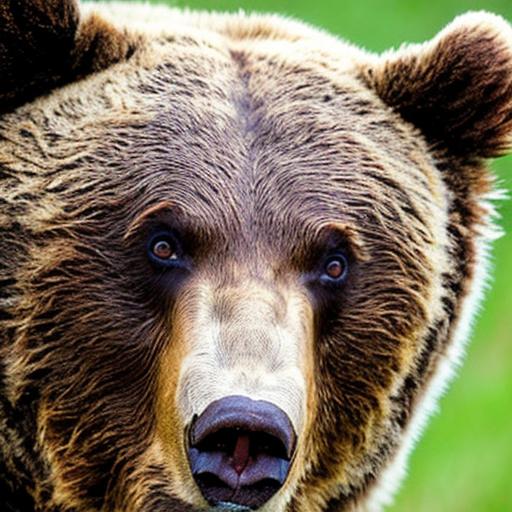}
        \\
        \includegraphics[scale=0.16]{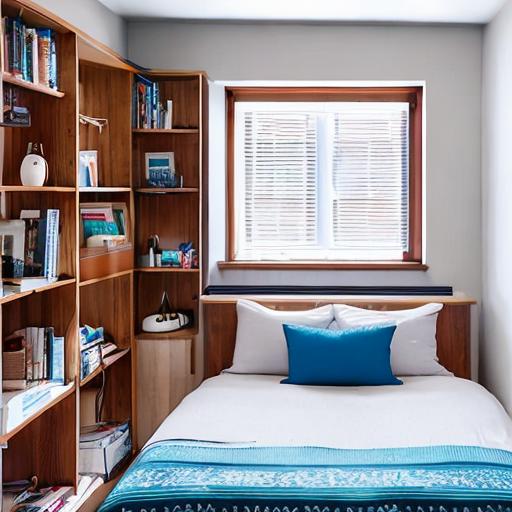} & 
        \includegraphics[scale=0.16]{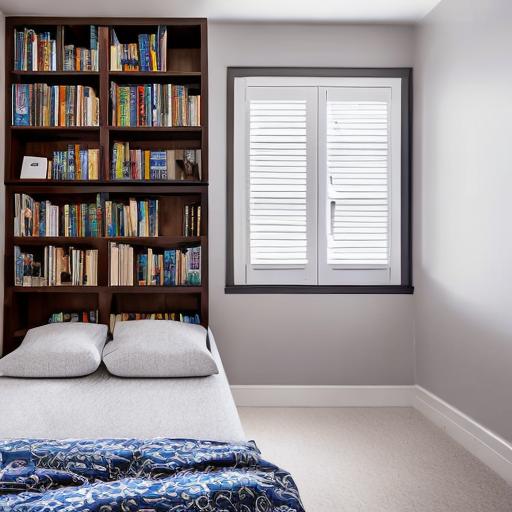} &
        \includegraphics[scale=0.16]{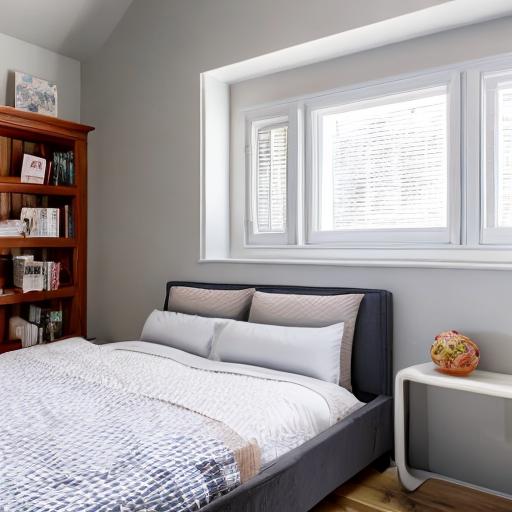} &
        \includegraphics[scale=0.16]{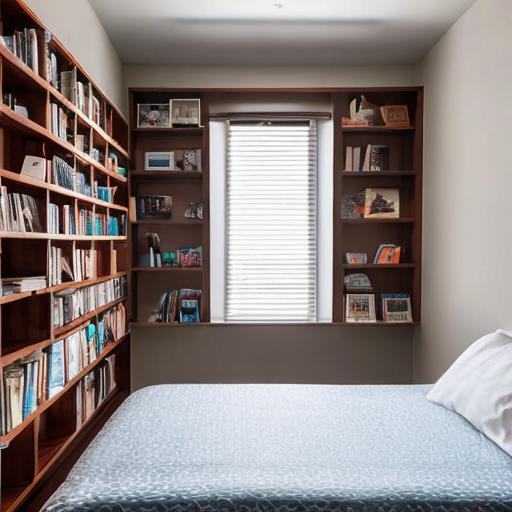}
        \\
        \includegraphics[scale=0.16]{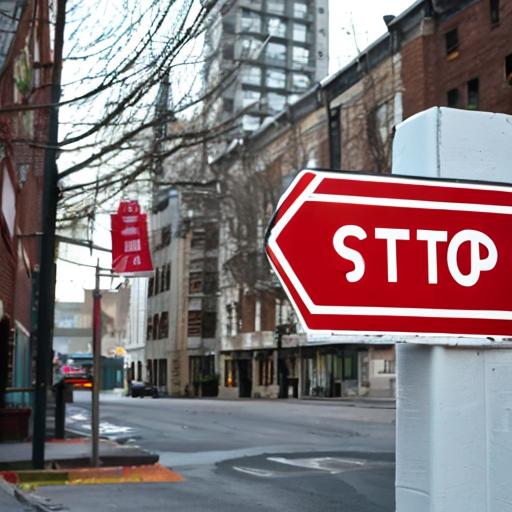} & 
        \includegraphics[scale=0.16]{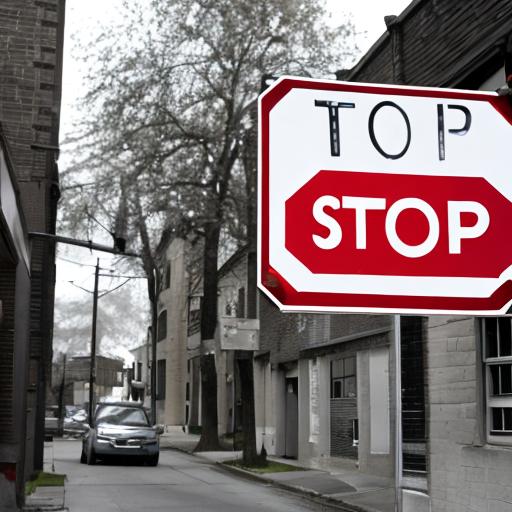} &
        \includegraphics[scale=0.16]{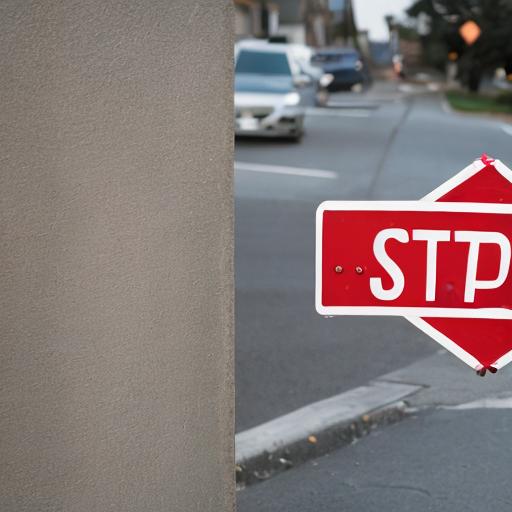} &
        \includegraphics[scale=0.16]{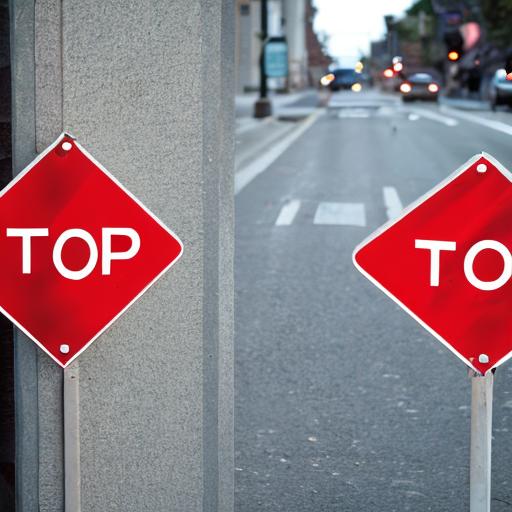}
        \\
        \includegraphics[scale=0.16]{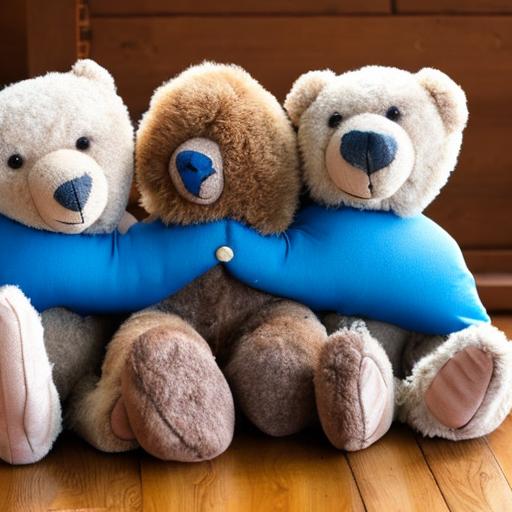} & 
        \includegraphics[scale=0.16]{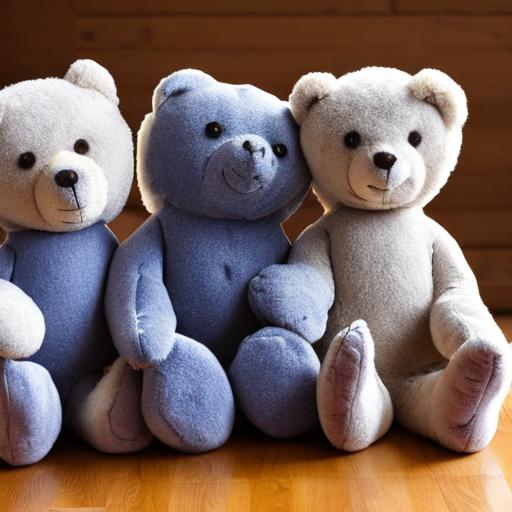} &
        \includegraphics[scale=0.16]{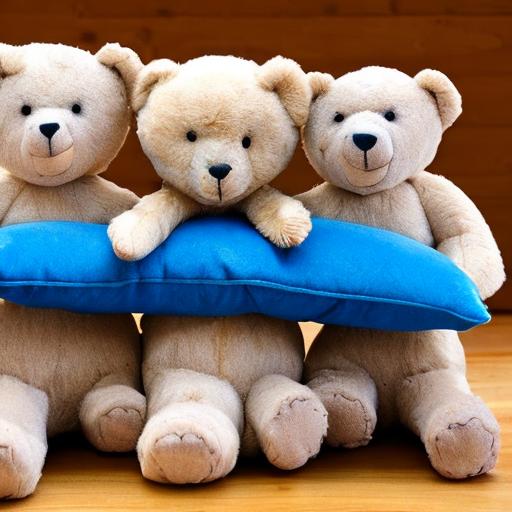} &
        \includegraphics[scale=0.16]{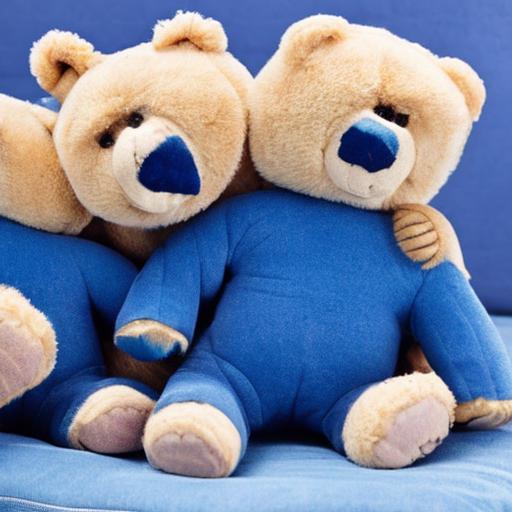}
        \\
        \includegraphics[scale=0.16]{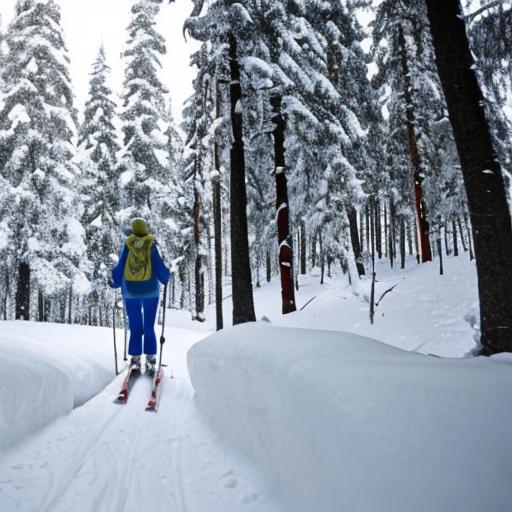} & 
        \includegraphics[scale=0.16]{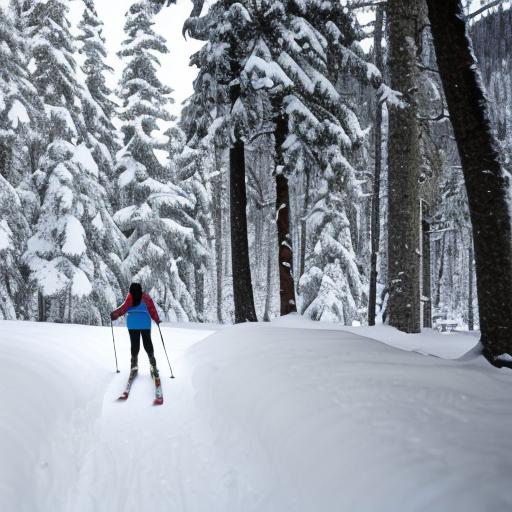} &
        \includegraphics[scale=0.16]{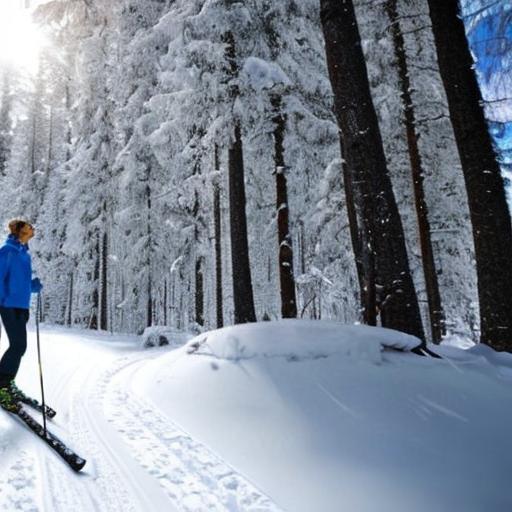} &
        \includegraphics[scale=0.16]{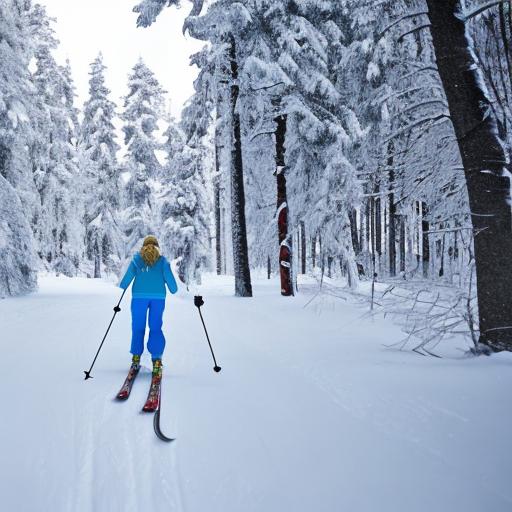}
        \\
        \includegraphics[scale=0.16]{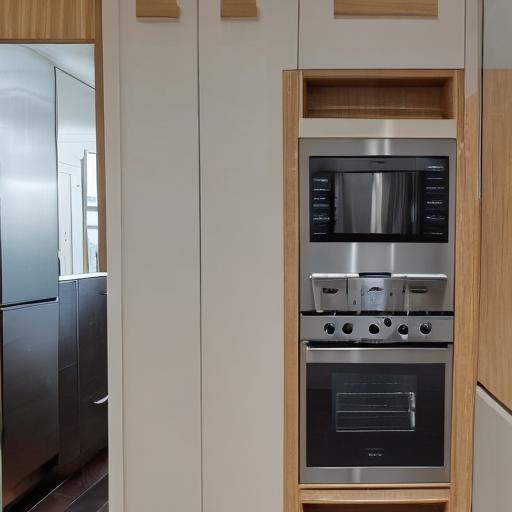} & 
        \includegraphics[scale=0.16]{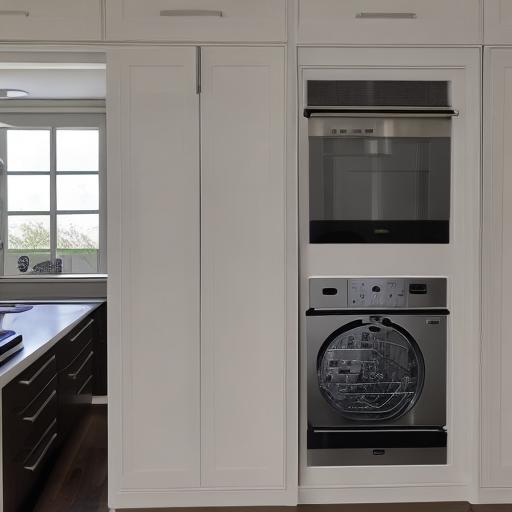} &
        \includegraphics[scale=0.16]{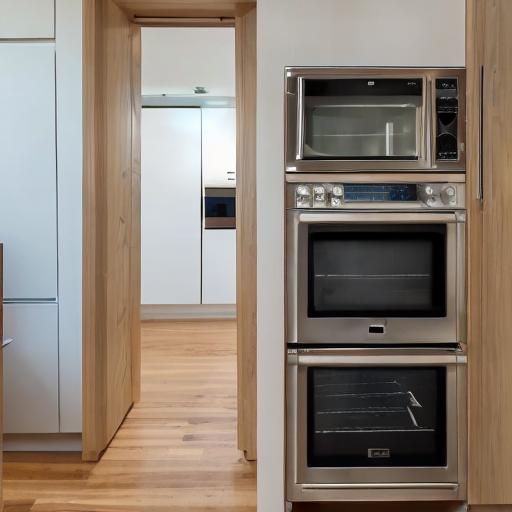} &
        \includegraphics[scale=0.16]{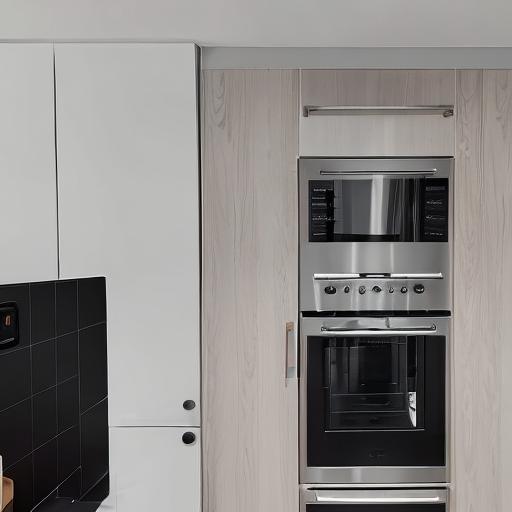}
    
    \end{tabular}
    \caption{More generated images with \textit{Tree-Ring Watermarking} with the first $7$ prompts in MS-COCO-2017 training dataset.}
    \label{fig:more_results}
\end{figure*}

\begin{figure}
    \centering
    \includegraphics[width=0.6\textwidth]{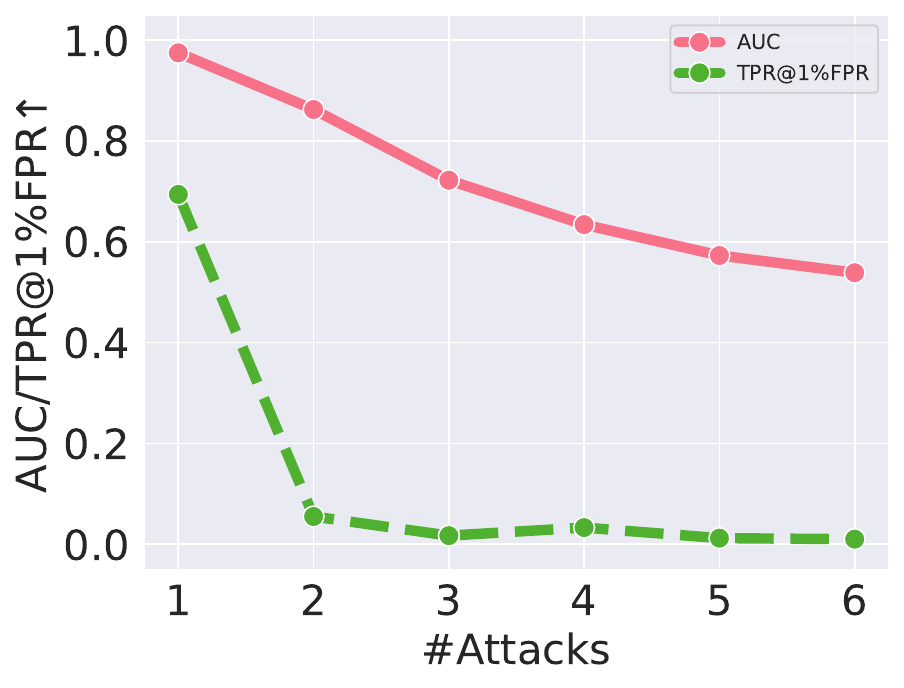}
    \caption{Results on $k$ number of random attacks applied at the same time.}
    \label{fig:abl-rand-attack}
\end{figure} 

\begin{figure}
    \centering
    \subfigure[$75^{\circ}$ Rotation]{\includegraphics[width=0.3\textwidth]{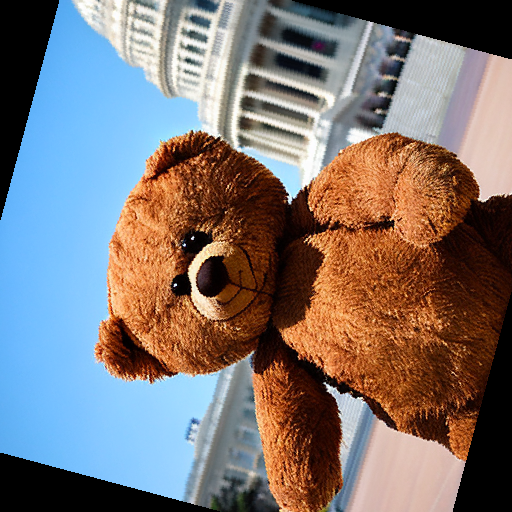}}
    \subfigure[$25\%$ JPEG Compression]{\includegraphics[width=0.3\textwidth]{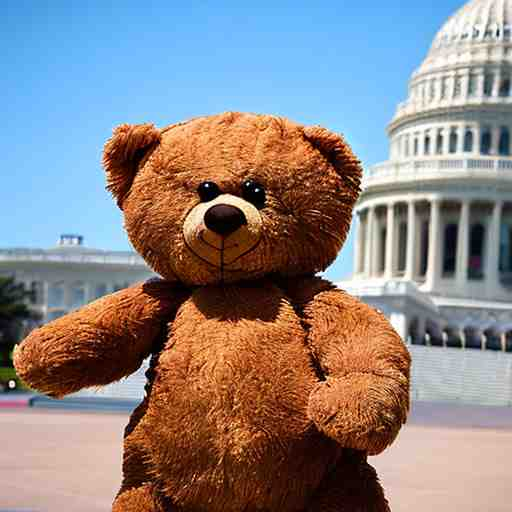}}
    \subfigure[$75\%$ Random Cropping + Scaling]{\includegraphics[width=0.3\textwidth]{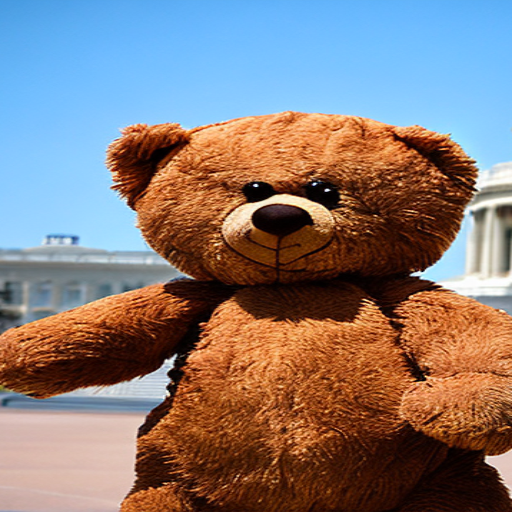}}
    \subfigure[Gaussian Blurring with an $8\times 8$ Filter]{\includegraphics[width=0.3\textwidth]{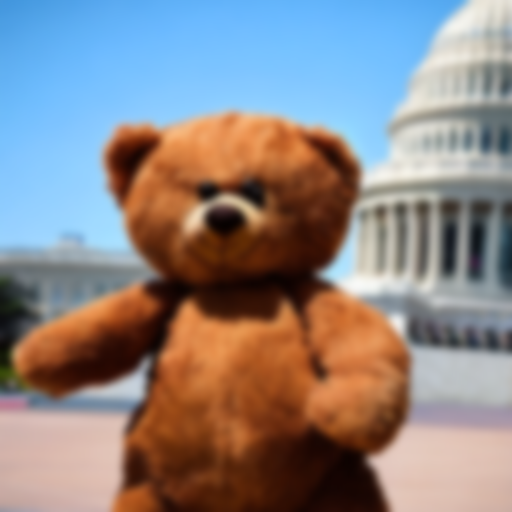}}
    \subfigure[Gaussian Noise with $\sigma=0.1$]{\includegraphics[width=0.3\textwidth]{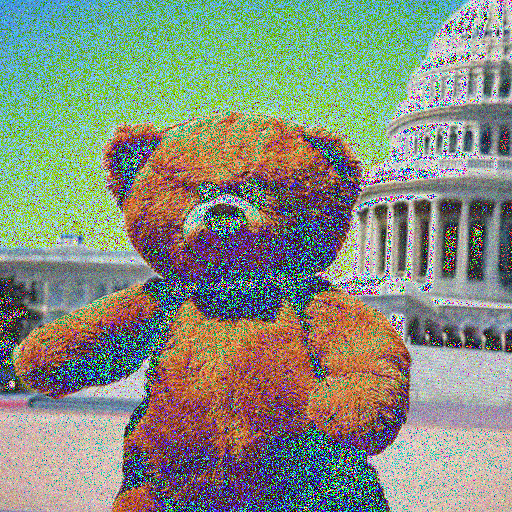}}
    \subfigure[Color Jitter with a Brightness Factor of $6$]{\includegraphics[width=0.3\textwidth]{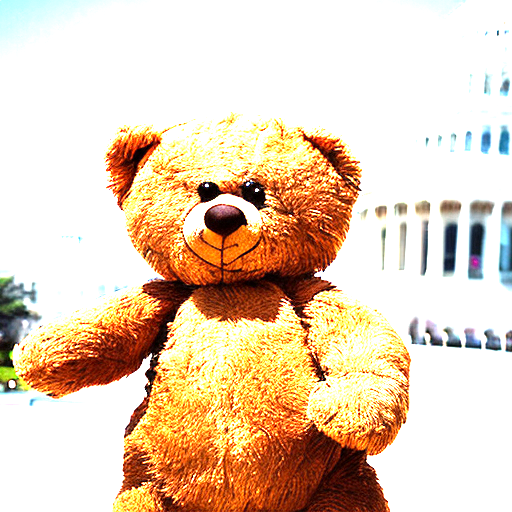}}
    \caption{Attacked images.}
    \label{fig:attack-viz}
\end{figure}

\end{document}